\documentclass[10pt,twocolumn,letterpaper]{article}

\usepackage{cvpr}
\usepackage{times}
\usepackage{epsfig}
\usepackage{graphicx}
\usepackage{amsmath}
\usepackage{amssymb}

\usepackage{amsmath,amsfonts,bm}

\def\eqref#1{equation~\ref{#1}}

\def\1{\bm{1}}

\DeclareMathAlphabet{\mathsfit}{\encodingdefault}{\sfdefault}{m}{sl}
\SetMathAlphabet{\mathsfit}{bold}{\encodingdefault}{\sfdefault}{bx}{n}

\DeclareMathOperator*{\argmin}{arg\,min}

\usepackage{booktabs} 
\usepackage{adjustbox}
\usepackage{multirow}
\usepackage{subfigure}
\usepackage{wrapfig}
\RequirePackage{algorithm}
\RequirePackage{algorithmic}

\usepackage[pagebackref=true,breaklinks=true,letterpaper=true,colorlinks,bookmarks=false]{hyperref}

\let\citep\cite
\let\citet\cite
\let\citealt\cite

\newcommand{\echo}[1]{{#1}}

\newcommand{\ms}[1]{\tiny{$\pm$#1}}
\newcommand{\eqr}[1]{(\ref{#1})}

\cvprfinalcopy 

\def\cvprPaperID{8365} 

\ifcvprfinal\fi
\begin{document}

\title{\emph{M2m}: Imbalanced Classification via Major-to-minor Translation}

\author{Jaehyung Kim\thanks{Equal contribution}
\qquad Jongheon Jeong\footnotemark[1]
\qquad Jinwoo Shin\\
Korea Advanced Institute of Science and Technology (KAIST)\\
Daejeon, South Korea\\
{\tt\small \{jaehyungkim,\,jongheonj,\,jinwoos\}@kaist.ac.kr}
}

\maketitle
\thispagestyle{empty}

\begin{abstract}
In most real-world scenarios, labeled training datasets are highly class-imbalanced, where deep neural networks suffer from generalizing to a balanced testing criterion.
In this paper, we explore a novel yet simple way to alleviate this issue by augmenting less-frequent classes via translating samples (e.g., images) from more-frequent classes.
This simple approach enables a classifier to learn more generalizable features of minority classes, by transferring and leveraging the diversity of the majority information. 
Our experimental results on a variety of class-imbalanced datasets show that the proposed method improves the generalization on minority classes significantly compared to other existing re-sampling or re-weighting methods. The performance of our method even surpasses 
those of previous state-of-the-art methods for the imbalanced classification.
\end{abstract}

\vspace{-0.05in}
\section{Introduction}
The recent success of deep neural networks (DNNs) across various computer vision problems \citep{he2016deep, Redmon_2017_CVPR, he2017mask, qi2017pointnet} has emerged due to the access to large-scale, annotated datasets collected from our visual world \cite{ILSVRC15,lin2014microsoft, abu2016youtube}. Despite having several well-organized datasets in research, \eg, CIFAR~\cite{dataset/cifar} and ILSVRC~\cite{ILSVRC15}, real-world datasets usually suffer from its expensive data acquisition process and the labeling cost. This commonly leads a dataset to have a ``long-tailed'' label distribution \citep{mahajan2018exploring, van2018inaturalist}. Such \emph{class-imbalanced} datasets make the standard training of DNN harder to generalize \citep{wang2017learning, ren2018learning, dong2018imbalanced}, particularly if one requires a class-balanced performance metric for a practical reason.

\begin{figure}[t]
\begin{center}
    \includegraphics[width=\linewidth]{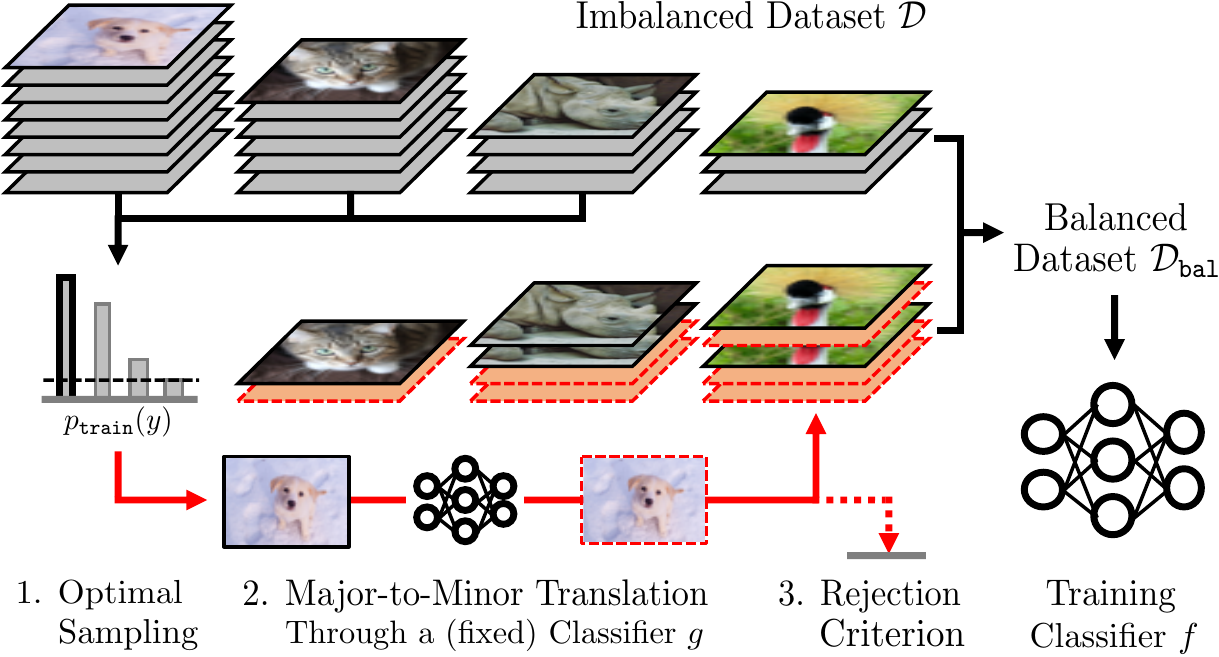}
\end{center}
\caption{An overview of the proposed method, called \emph{Major-to-minor Translation} (M2m). M2m is based on the over-sampling method, and attempts to replace the over-sampled (duplicated) minority samples with synthetic ones translated from other majority samples. The more details are presented in Section~\ref{s:M2m}.} 
\vspace{-0.05in}
\label{fig:ams}
\vspace{-0.15in}
\end{figure}

A natural approach in an attempt to bypass this \emph{class-imbalance problem} is to re-balance the training objective artificially with respect to the class-wise sample sizes. Two of such methods are representative: (a) \emph{re-weighting} the given loss function by a factor inversely proportional to the sample frequency in a class-wise manner \citep{huang2016learning, khan2017cost}, and (b) \emph{re-sampling} the given dataset so that the expected sampling distribution during training can be balanced, either by ``over-sampling'' the minority classes \citep{japkowicz2000class, cui2018large} or ``under-sampling'' the majority classes \citep{he2008learning}.

However, na\"ively re-balancing the objective usually results in harsh over-fitting to minority classes, since they cannot handle, the lack of minority information in essence. Several attempts have been made to alleviate this issue: Cui~\etal\citet{cui2019class} proposed the concept of ``effective number'' of samples as alternative weights in the re-weighting method. Cao~\etal \citet{cao2019learning} found that both re-weighting and re-sampling can be much more effective when applied at the later stage of training, in case of neural networks. In the context of re-sampling, SMOTE \citep{chawla2002smote} is a widely-used variant of the over-sampling method that mitigates the over-fitting via data augmentation, and several variants of SMOTE have been suggested accordingly \cite{han2005borderline, he2008adasyn, Mullick_2019_ICCV}. A major drawback of these SMOTE-based methods is that they usually perform poorly when there exist only a few samples in the minority classes, \ie, under regime of ``extreme'' imbalance, because they synthesize a new minority sample only using the existing samples of the same class.

Another line of research attempts to prevent the over-fitting with a new regularization scheme that minority classes are more penalized, where the margin-based approaches generally suit well as a form of data-dependent regularizer \citep{zhang2017range, dong2018imbalanced, khan2019striking, cao2019learning}. There have also been works that view the class-imbalance problem in the framework of active learning \citep{ertekin2007learning, attenberg2013class} or meta-learning \citep{wang2017learning, ren2018learning, shu2019meta, liu2019large}.

\vspace{-0.15in}
\paragraph{Contribution.} 
In this paper, we revisit the over-sampling framework and propose a new way of generating minority samples, coined \emph{Major-to-minor Translation} (M2m). In contrast to other over-sampling methods, \eg, SMOTE that applies data augmentation to minority samples to mitigate the over-fitting issue, we attempt to generate minority samples in a completely different way. The proposed M2m does \emph{not} use the existing minority samples for the over-sampling. Instead, it use the \emph{majority} (non-minority) samples and translate them to the target minority class using another classifier independently trained under the given imbalanced dataset. Our key finding is that, this method turns out to be very effective on learning more generalizable features in imbalanced learning: it does not overly use the minority samples, and leverages the richer information of the majority samples simultaneously. 

Our minority over-sampling method consists of three components to improve the sampling quality. First, we propose an optimization objective for generating synthetic samples: a majority sample can be translated into a synthetic minority sample via optimizing it, while not affecting the performance of the majority class (even the sample is labeled to the minority class). Second, we design a sample rejection criterion based on the observation that generation from more majority class is more preferable. Third, based on the proposed rejection criterion, we suggest an optimal distribution for sampling a majority seed to be translated in our generation process.

We evaluate our method on various imbalanced classification problems, covering synthetically imbalanced datasets from CIFAR-10/100 \citep{dataset/cifar} and ImageNet \citep{liu2019large}, and real-world imbalanced datasets including {CelebA \cite{liu2015faceattributes}}, SUN397 \cite{xiao2010sun}, Twitter \citep{gimpel2010part} and Reuters \citep{lewis2004rcv1} datasets. Despite its simplicity, our method significantly improves the balanced test accuracy compared to previous re-sampling or re-weighting methods across all the tested datasets. Our results even surpass those from LDAM~\citealt{cao2019learning}, a current state-of-art margin-based method. Moreover, we found our method is particularly effective under ``extreme'' imbalance: in the case of Reuters of the most severe imbalance, we could improve the balanced accuracy by (relatively) ${17.1\%}$ and ${9.2\%}$ upon standard training and LDAM, respectively.

\section{\emph{M2m}: Major-to-minor translation}
\label{s:M2m}
We consider a classification problem with $K$ classes from a dataset $\mathcal{D}=\{(x_i, y_i)\}_{i=1}^N$, where $x\in\mathbb{R}^d$ and $y\in\{1,\cdots,K\}$ denote an input and the corresponding class label, respectively. Let $f: \mathbb{R}^d \rightarrow \mathbb{R}^K$ be a classifier designed to output $K$ logits, which we want to train against the class-imbalanced dataset $\mathcal{D}$. We let $N:=\sum_{k}N_k$ denote the total sample size of $\mathcal{D}$, where $N_k$ is that of class $k$. Without loss of generality, we assume $N_1 \ge N_2 \ge \cdots \ge N_K$. In the \emph{class-imbalanced} classification, the class-conditional data distributions $\mathcal{P}_k := p(x~|~y=k)$ are assumed to be invariant across training and test time, but they have different prior distributions, say $p_{\tt train}(y)$ and $p_{\tt test}(y)$, respectively: $p_{\tt train}(y)$ is highly imbalanced while $p_{\tt test}(y)$ is usually assumed to be the uniform distribution. The primary goal of the class-imbalanced learning is to train $f$ from $\mathcal{D}\sim\mathcal{P}_{\tt train}$ that generalizes well under $\mathcal{P}_{\tt test}$ compared to the standard training, \eg, empirical risk minimization (ERM) with an appropriate loss function $\mathcal{L}(f)$:
\begin{equation}\label{eq:erm}
\min_{f}~{\mathbb{E}_{(x, y)\sim\mathcal{D}}[\mathcal{L}(f; x, y)]}.
\end{equation}
Our method is primarily based on over-sampling technique \citep{japkowicz2000class}, a traditional and principled way to balance the class-imbalanced training objective via sampling minority classes more frequently. In other words, we assume a ``virtually balanced'' training dataset $\mathcal{D}_{\tt bal}$ made from $\mathcal{D}$ such that the class $k$ has $N_1 - N_k$ more samples, and the classifier $f$ is trained on $\mathcal{D}_{\tt bal}$ instead of $\mathcal{D}$. 

A key challenge in over-sampling is to prevent \emph{over-fitting} on minority classes, as the objective modified is essentially much biased to a few samples of minority classes. In contrast to most prior works that focus on performing data augmentation \emph{directly} on minority samples to mitigate this issue \citep{chawla2002smote, liu2019large, Mullick_2019_ICCV}, we attempt to augment minority samples in a completely different way: our method does \emph{not} use the minority samples for the augmentation, but the majority samples.

\subsection{Overview of M2m}
\label{ss:overview}

Consider a scenario of training a neural network $f$ on a class-imbalanced dataset $\mathcal{D}$. The proposed \emph{Major-to-minor Translation} (M2m) attempts to construct a new balanced dataset $\mathcal{D}_{\tt bal}$ for training $f$, by adding synthetic minority samples that are \emph{translated} from other samples of (relatively) majority classes. There could be multiple ways to perform this ``Major-to-minor'' translation. In particular, a recent progress on cross-domain generation via generative adversarial networks \cite{zhu2017unpaired, choi2018stargan, mo2018instagan} has made this more attractive, provided that much computational cost for additional training is acceptable. In this paper, on the other hand, we explore a much simpler and efficient approach: we translate a majority sample by optimizing it to \emph{maximize} the target minority confidence of another baseline classifier $g$. Here, we assume the classifier $g$ is a pre-trained neural network on $\mathcal{D}$ so that performs well (at least) on the training imbalanced dataset, \eg, via standard ERM training. This implies that, $g$ may be over-fitted to minority classes and does not necessarily generalize well under the balanced test dataset. We found this mild assumption on $g$ is fairly enough to capture the information in the small minority classes and could generate surprisingly useful synthetic minority samples by utilizing the diversity of majority samples. On the other hand, $f$ is the target network that we aim to train to perform well on the balanced testing criterion.

\begin{figure}[t]
\begin{center}
    \includegraphics[width=0.52\linewidth]{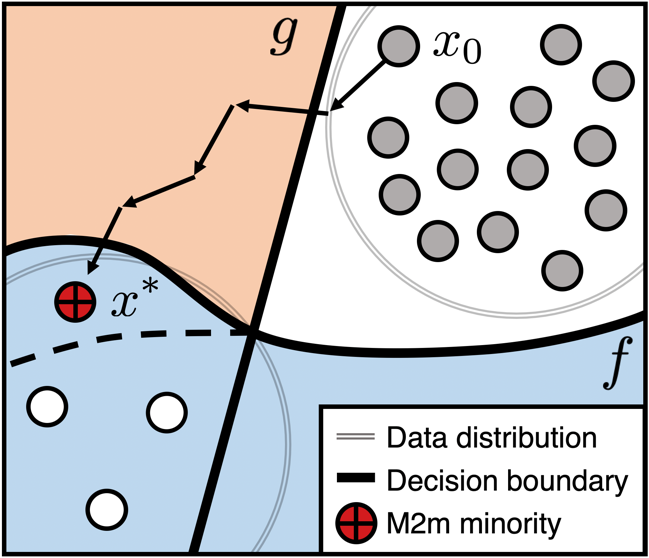}
\end{center}
\caption{An illustration of M2m generation. A majority seed $x_{0}$ is translated to a synthetic minority $x^{*}$ based on the decision boundary of $g$.
By incorporating $x^*$, $f$ learns an extended decision boundary of the target minority class.} \label{fig:m2m_opt}
\end{figure}

During the training of $f$, M2m utilizes the given classifier $g$ to generate new minority samples, and the generated samples are added to $\mathcal{D}$ to construct $\mathcal{D}_{\tt bal}$ on the fly. To obtain a single synthetic minority $x^{*}$ of class $k$, our method solves an optimization problem starting from another training sample $x_0$ of a (relatively) major class $k_0 < k$:
\begin{equation}\label{eq:obj}
x^{*} = \argmin_{x := x_0 + \delta}\ \mathcal{L}(g; x, k) + \lambda \cdot f_{k_0}(x),
\end{equation}
where $\mathcal{L}$ denotes the cross entropy loss and $\lambda>0$ is a hyperparameter.
In other words, our method ``translates'' a majority seed $x_0$ into $x^{*}$, so that $g$ confidently classifies it as minority class $k$. The generated sample $x^*$ is then labeled to class $k$ and fed into $f$ for training to perform better on $\mathcal{D}_{\tt bal}$ and match the prediction of $f$ to that of $g$. We do not force $f$ in \eqr{eq:obj} to classify $x^*$ to class $k$ as well, but we restrict that $f$ to have lower confidence on the original class $k_0$ by imposing a regularization term $\lambda \cdot f_{k_0}(x)$. Here, the regularization term $\lambda \cdot f_{k_0}(x)$ on the logit reduces the risk when $x^*$ is labeled to $k$, whereas it may contain significant features of $x_0$ in the viewpoint of $f$. Intuitively, one can regard the overall process as teaching $f$ to learn novel minority features which $g$ considers it significant, \ie, via extension of the decision boundary from the knowledge $g$.  Figure~\ref{fig:m2m_opt} illustrates the basic idea of our method.

\subsection{Underlying intuition on M2m}
\label{ss:intuition}

One may understand our method better by considering the case when $g$ is an ``oracle'' (possibly the Bayes optimal) classifier, \eg, (roughly) humans. Here, solving \eqr{eq:obj} essentially requires a transition of the original input $x_0$ of class $k_0$ with $100\%$ confidence to another class $k$ with respect to $g$: this would let $g$ ``erase and add'' the features related to the class $k_0$ and $k$, respectively. Hence, in this case, our process corresponds to collecting more in-distribution minority data, which may be argued as the best way one could do to resolve the class-imbalance problem.

An intriguing point here is, however, that neural network models are very far from this ideal behavior, even when they achieve super-human performance. Instead, when $f$ and $g$ are neural networks, \eqr{eq:obj} often finds $x^*$ that is very close to $x_0$, \ie, similar to the phenomenon of \emph{adversarial examples} \citep{szegedy2013intriguing, goodfellow2014explaining}. Nevertheless, we found our method still effectively improves the generalization of minority classes even in such cases. This observation is, in some sense, aligned to a recent claim that adversarial perturbation is not a ``bug'' in neural networks, but a ``generalizable'' feature \citep{ilyas2019adversarial}. 

In this paper, we hypothesize this counter-intuitive effectiveness of our method comes from mainly in two aspects: (a) the sample diversity in the majority dataset is utilized to prevent over-fitting on the minority classes, and (b) another classifier $g$ is enough to capture the information in the small minority dataset. In this respect, adversarial examples from a majority to a minority can be regarded as one of natural ways to leverage the diverse features in majority examples useful to improve the generalization of the minority classes. It is also notable that our over-sampling method does not completely replace the existing dataset. Instead, our method only \emph{augment} the minority classes, and our finding is that this augmentation turns out to be very effective than na\"ively duplicating minority examples as done by the standard over-sampling. We further discuss a more detailed analysis to verify these claims, by performing an extensive ablation study in Section~{\ref{ss:ablation}}.

\subsection{Detailed components of M2m}
\paragraph{Sample rejection criterion.}
An important factor that affects the quality of the synthetic minority samples in our method is the quality of $g$, especially for $g_{k_0}$: a better $g_{k_0}$ would more effectively ``erase'' important features of $x_0$ during the translation, thereby making the resulting minority samples more reliable. In practice, however, $g$ is not that perfect: the synthetic samples still contain some discriminative features of the original class $k_0$, in which it may even harm the performance of $f$. This risk of ``unreliable'' generation becomes more harsh when $N_{k_0}$ is small, as we assume that $g$ is also trained on the given imbalanced dataset $\mathcal{D}$.

To alleviate this risk, we consider a simple criterion for \emph{rejecting} each of the synthetic samples randomly with probability depending on $k_0$ and $k$:
\begin{equation}\label{eq:rej}
\mathbb{P}(\text{Reject } x^*| k_0, k) := \beta^{(N_{k_0}-N_k)^+},
\end{equation}
where $(\cdot)^+ := \max(\cdot, 0)$, and $\beta \in [0, 1)$ is a hyperparameter which controls the reliability of $g$: the smaller $\beta$, the more reliable $g$. For example, if $\beta=0.999$, the synthetic samples are accepted with probability more than $99\%$ if $N_{k_0} - N_k > 4602$. When $\beta=0.9999$, on the other hand, it requires $N_{k_0} - N_k > 46049$ to achieve the same goal. This exponential modeling of the rejection probability is motivated by the \emph{effective number} of samples \citep{cui2019class}, a heuristic recently proposed to model the observation that the impact of adding a single data point exponentially decreases at larger datasets. When a synthetic sample is rejected, we simply replace it by an existing minority sample from the original dataset $\mathcal{D}$ to obtain the balanced dataset $\mathcal D_{\tt bal}$.

\renewcommand{\algorithmicrequire}{\textbf{Input:}}
\renewcommand{\algorithmicensure}{\textbf{Output:}}

\begin{algorithm}[t]
\caption{Over-sampling via M2m}\label{alg:training}
\begin{algorithmic}[1]
\REQUIRE A dataset $\mathcal{D}=\{(x_{i}, y_{i})\}_{i=1}^N$ with $N=\sum_{k=1}^K N_k$. A classifier $f$. A pre-trained classifier $g$. $\lambda, \gamma, \eta, T > 0$ and $\beta\in[0, 1)$.
\ENSURE A class-balanced dataset $\mathcal{D}_{\tt bal}$
\vspace{0.05in}
\hrule
\vspace{0.05in}
\STATE Initialize $\mathcal{D}_{\tt bal} \leftarrow \mathcal{D}$
\FOR{$k=2$ {\bfseries to} $K$}
    \STATE $\Delta \leftarrow N_1 - N_k$ 
    \FOR{$i=1$ {\bfseries to} $\Delta$}
        \STATE $k_0 \sim Q(k_0|k) \propto 1 - \beta^{(N_{k_0}-N_k)^+}$
        \STATE $x_0 \leftarrow$ A random sample of class $k_0$ in $\mathcal{D}$  
        \STATE Initialize $x^{*} \leftarrow x_0 + \delta$ with a small noise $\delta$
        \FOR{$t=1$ {\bfseries to} $T$}
            \STATE $\delta \leftarrow \nabla_{x^{*}}[\mathcal{L}(g; x^{*}, k) + \lambda \cdot f_{k_0}(x^{*})]$
            \STATE $x^{*} \leftarrow x^{*} - \eta\cdot\frac{\delta}{||\delta||_2}$
        \ENDFOR
        \STATE $R \sim \mathrm{Bernoulli}(\beta^{(N_{k_0}-N_k)^+})$ 
        \IF{$\mathcal{L}(g; x^*, k) > \gamma$ \OR $R = 1$}
        \STATE $x^* \leftarrow$ A random sample of class $k$ in $\mathcal{D}$
        \ENDIF
    \STATE $\mathcal{D}_{\tt bal} \leftarrow \mathcal{D}_{\tt bal} \cup \{(x^*, k)\}$
    \ENDFOR
\ENDFOR
\end{algorithmic}
\end{algorithm}
\vspace{-0.12in}
\paragraph{Optimal seed sampling.}
Another design choice of our method is \emph{how to choose} a (majority) seed sample $x_0$ with class $k_0$ for each generation in \eqr{eq:obj}. Based on the rejection criterion proposed in \eqr{eq:rej}, we design a sampling distribution $Q(k_0| k)$ for selecting the class $k_0$ of initial point $x_0$ given target class $k$, by considering two aspects: (a) $Q$ maximizes the \emph{acceptance probability} $P_{\tt accept}(k_0| k)$ under our rejection criterion, and (b) $Q$ chooses \emph{diverse} classes as much as possible, \ie, the entropy $H(Q)$ is maximized.  Namely, we are interested in the following optimization:
\begin{equation}\label{eq:kldiv}
\max_{Q}{\bigg[\underbrace{\mathbb{E}_Q [\log P_{\tt accept}]}_{\text{(a)}} + \underbrace{H(Q)}_{\text{(b)}} \bigg]}.
\end{equation}
It is elementary to check that $Q=P_{\tt accept}$ is the solution of the above optimization. Hence, due to the rejection probability \eqr{eq:rej}, we choose: 
\begin{equation}\label{eq:acc}
Q(k_0| k) \propto 1 - \beta^{(N_{k_0}-N_k)^+}.
\end{equation}
Once $k_0$ is selected, a sample $x_0$ is sampled uniformly at random among samples having the class $k_0$. The overall procedure of M2m is summarized in Algorithm~\ref{alg:training}.

\vspace{-0.2in}
\paragraph{Practical implementation via re-sampling.}
In practice of training a neural network $f$, \eg, stochastic gradient descent (SGD) with a mini-batch sampling, M2m is implemented using a batch-wise re-sampling. More precisely, in order to simulate the generation of $N_1 - N_k$ samples for any $k=1, 2, \cdots, K$, we perform the generation with probability $\frac{N_1 - N_{y_i}}{N_1}=1 - N_{y_i}/N_1$, for all $i$ in a given {class-balanced} mini-batch $\mathcal{B}=\{(x_i, y_i)\}_{i=1}^m$.\footnote{Obtaining such a class-balanced mini-batch can be done via standard re-sampling.} For a single generation at index $i$, we first sample $k_0\sim Q(k_0|y_i)$ following \eqr{eq:acc} until $k_0\in\{y_i\}_{i=1}^m$, and select a seed $x_0$ of class $k_0$ randomly inside $\mathcal{B}$.  Then, we solve the optimization \eqr{eq:obj} from $x_0$ toward class $y_i$ via gradient descent for a fixed number of iterations $T$ with a step size $\eta$. We  accept the result sample $x^*$ only if $\mathcal{L}(g; x^*, y_i)$ is less than $\gamma>0$ for stability. Finally, if accepted, we replace $(x_i, y_i)$ in $\mathcal{B}$ by $(x^*, y_i)$.

\section{Experiments}
\begin{figure*}[t]
\begin{center}
    {
    \subfigure[CIFAR-LT-10]
    {
    \includegraphics[width=0.15\textwidth]{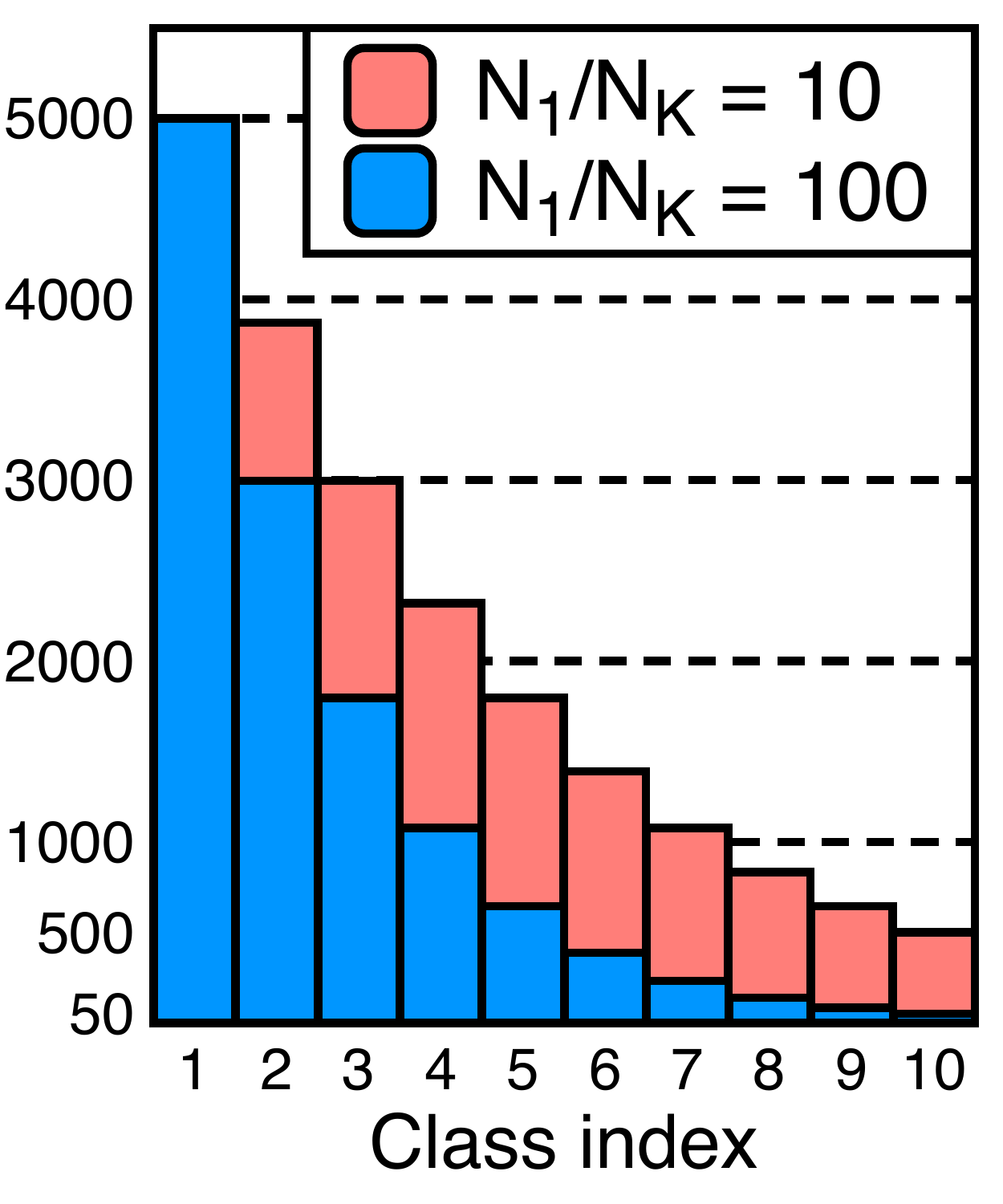}
    \label{fig:CIFAR-10}}
    \subfigure[CIFAR-LT-100]
    {
    \includegraphics[width=0.15\textwidth]{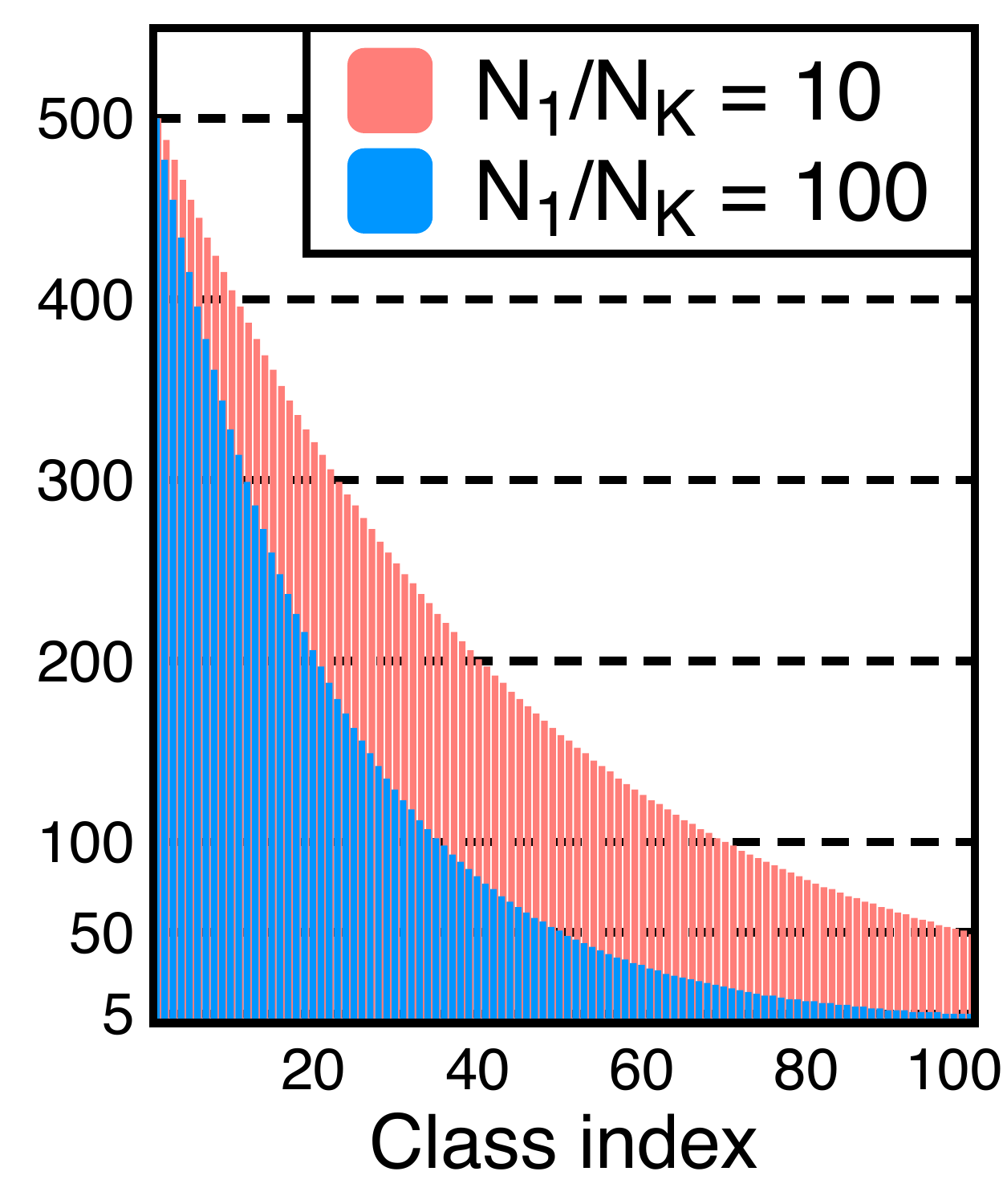}
    \label{fig:CIFAR-100}} 
    \subfigure[CelebA-5]
    {
    \includegraphics[width=0.15\textwidth]{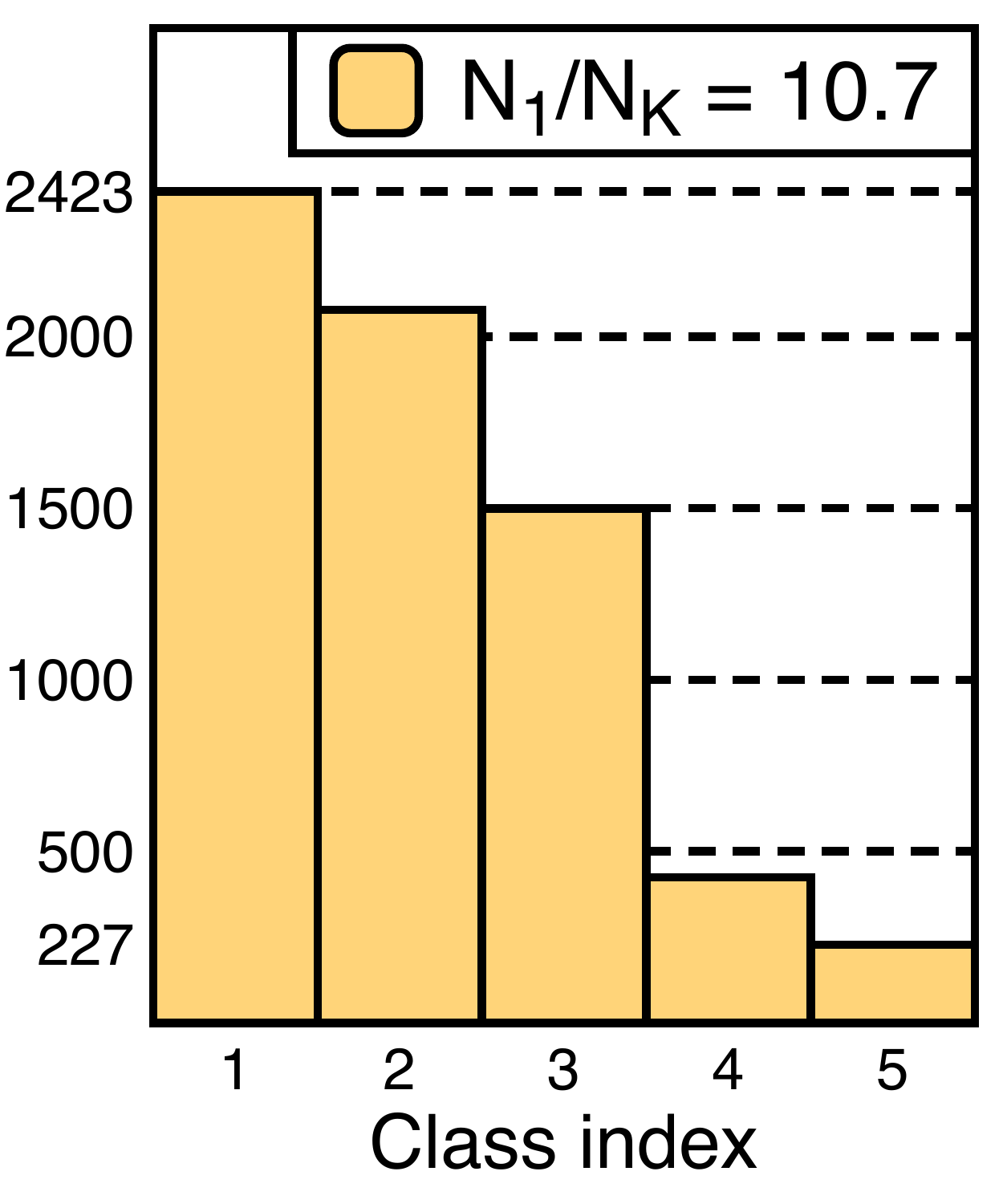}
    \label{fig:celeba}
    }
    \subfigure[SUN397]
    {
    \includegraphics[width=0.15\textwidth]{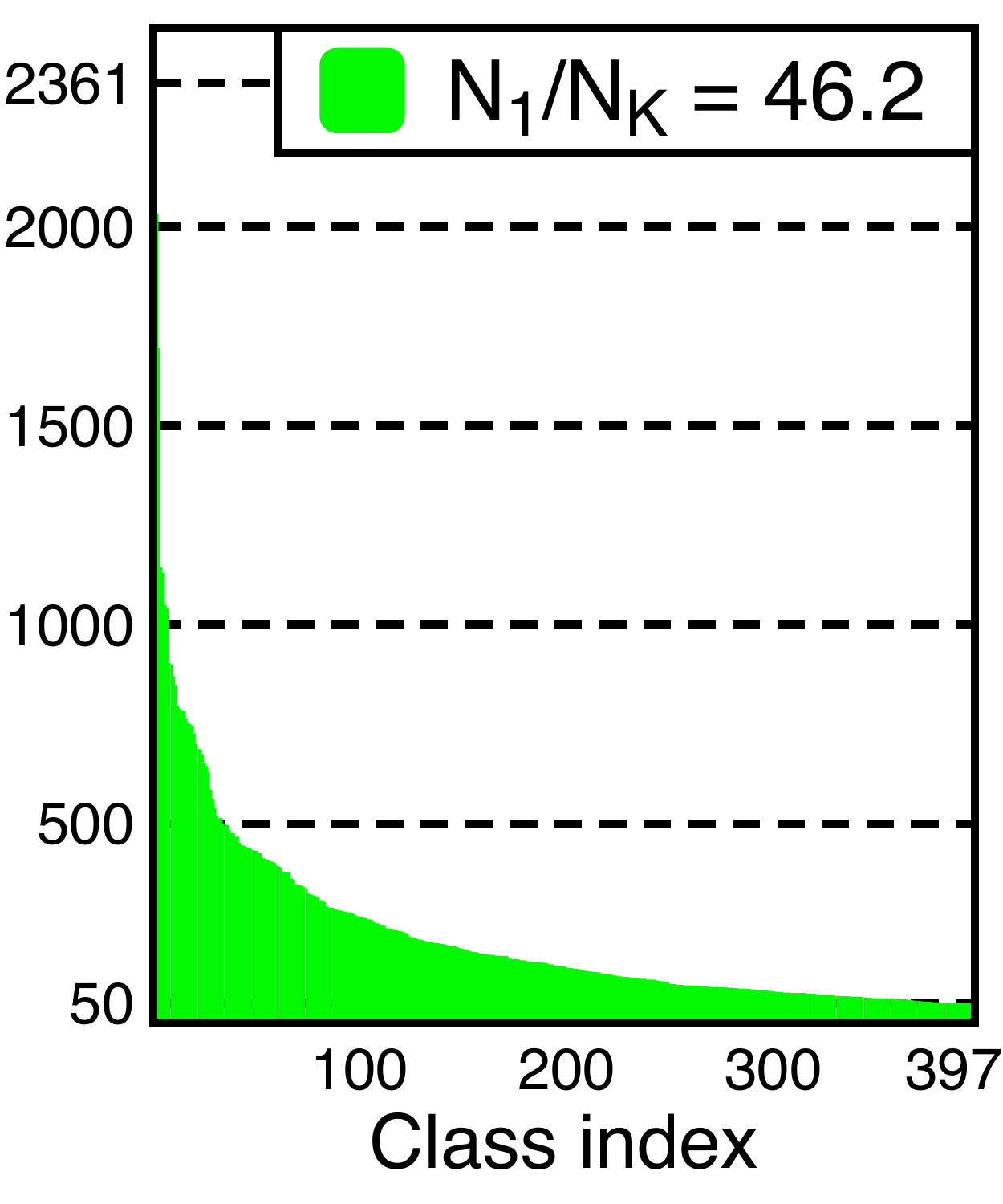}
    \label{fig:Sun}
    }
    \subfigure[Twitter]
    {
    \includegraphics[width=0.15\textwidth]{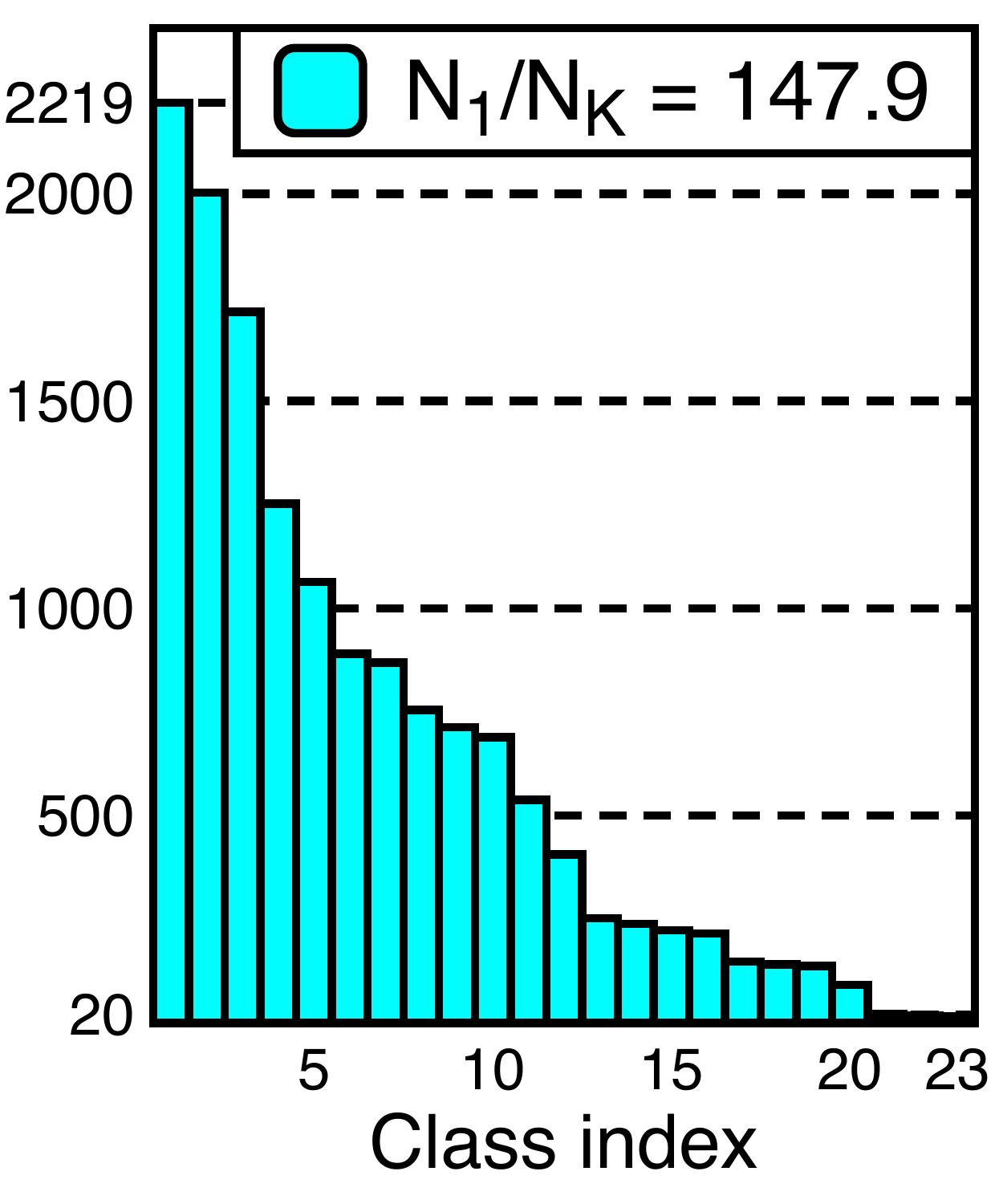}
    \label{fig:Twitter}}
    \subfigure[Reuters]
    {
    \includegraphics[width=0.15\textwidth]{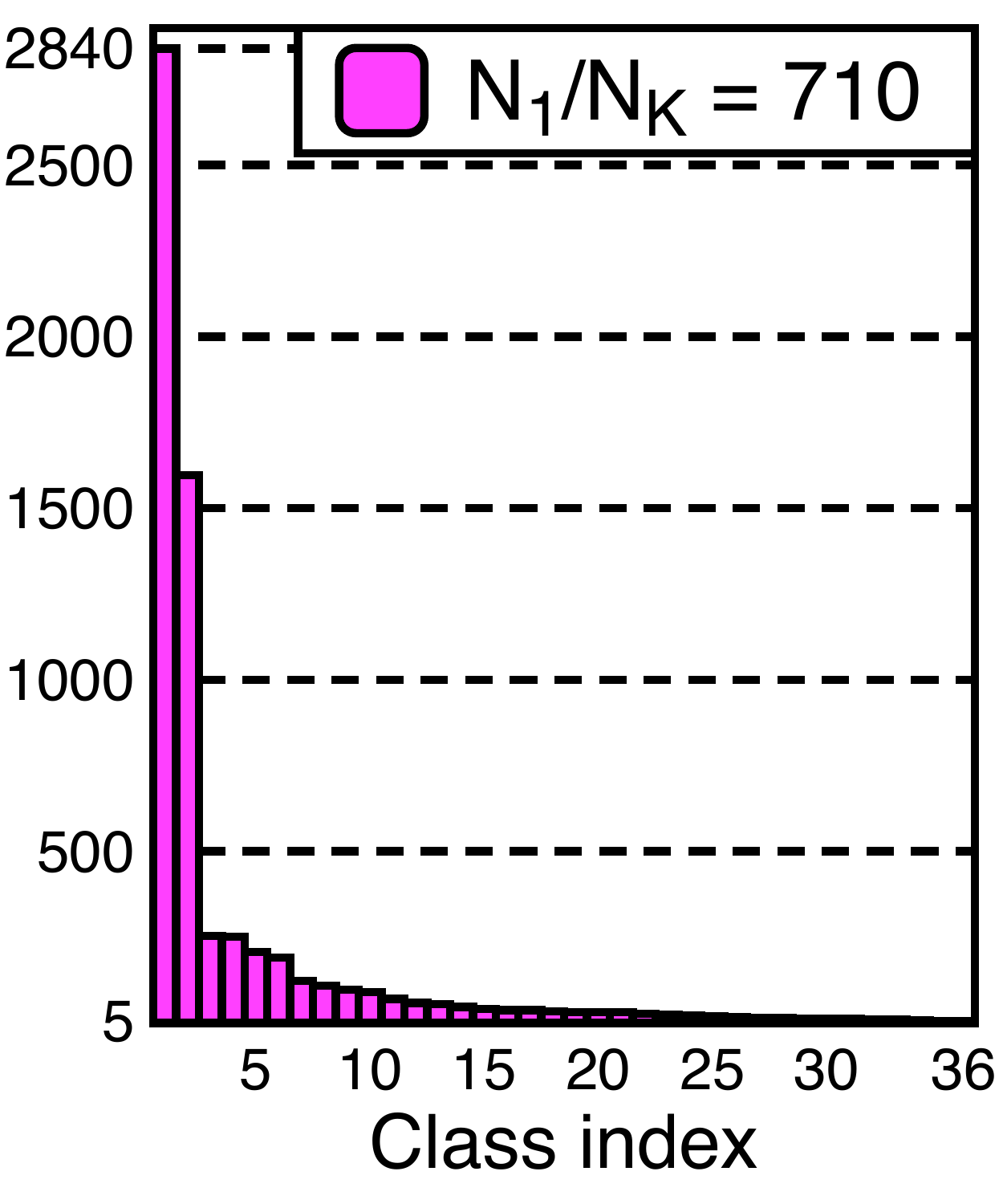}
    \label{fig:Reuter}}
    }
\end{center}
\vspace{-0.05in}
\caption{An illustration of histograms on training sample sizes for the datasets used in this paper.}
\label{fig:stat}
\vspace{-0.1in}
\end{figure*}

We evaluate our method on various class-imbalanced classification tasks:
synthetically-imbalanced variants of CIFAR-10/100~\citep{dataset/cifar},
ImageNet-LT\footnote{Results on ImageNet-LT can be found in the supplementary material.}~\citep{liu2019large}, CelebA~\cite{liu2015faceattributes}, SUN397~\cite{xiao2010sun}, Twitter~\citep{gimpel2010part}, and Reuters~\citep{lewis2004rcv1} datasets.\footnote{Code is available at \url{https://github.com/alinlab/M2m}} Figure~\ref{fig:stat} illustrates the class-wise sample distributions for the datasets considered in our experiments.
The more details on the tested datasets are given in the supplementary material. To evaluate the classification performance of the models on the balanced test distribution, we mainly report two popular metrics: the \emph{balanced accuracy} (bACC) \citealt{huang2016learning, wang2017learning} and the \emph{geometric mean scores} (GM) \citealt{kubat1997addressing, branco2016survey}, which are defined by the arithmetic and geometric mean over class-wise sensitivity (\ie, recall), respectively. 
We remark that bACC is essentially equivalent to the standard accuracy metric for balanced datasets. All the values and error bars in this section are mean and standard deviation across three random trials, respectively.
Overall, our results clearly demonstrate that minority synthesis via translating from majority consistently improves the efficiency of over-sampling, in terms of the significant improvement of the generalization in minority classes compared to other re-sampling baselines, across all the tested datasets. We also perform an ablation study to verify the effectiveness of our main ideas. 

\subsection{Experimental setup}
\label{ss:setup}

\paragraph{Baseline methods.}
We consider a wide range of baseline methods, as listed in what follows: 
(\textit{a}) empirical risk minimization (ERM): training on the cross-entropy loss without any re-balancing; 
(\textit{b}) re-sampling (RS) \citealt{japkowicz2000class}: balancing the objective from different sampling probability for each sample; 
(\textit{c}) SMOTE \citep{chawla2002smote}: a variant of re-sampling with data augmentation;
(\textit{d}) re-weighting (RW) \citealt{huang2016learning}: balancing the objective from different weights on the sample-wise loss;
(\textit{e}) class-balanced re-weighting (CB-RW) \citealt{cui2019class}: a variant of re-weighting that uses the inverse of effective number for each class, defined as \mbox{$(1-\beta^{N_k})/(1-\beta)$}. Here, we use $\beta=0.9999$;
(\textit{f}) deferred re-sampling (DRS) \citealt{cao2019learning} and (\textit{g}) deferred re-weighting (DRW) \citealt{cao2019learning}: re-sampling and re-weighting is deferred until the later stage of the training, repsectively;
(\textit{h}) focal loss (Focal) \citealt{lin2017focal}: the objective is up-weighted for relatively hard examples to focus more on the minority; 
(\textit{i}) label-distribution-aware margin (LDAM) \citealt{cao2019learning}: the classifier is trained to impose larger margin to minority classes. 
Roughly, the considered baselines can be classified into three categories: (i) ``re-sampling'' based methods - (\textit{b, c, f}), (ii) ``re-weighting'' based methods - (\textit{d, e, g}), and (iii) different loss functions - (\textit{a, h, i}). 

\vspace{-0.15in}
\paragraph{Training details.}
We train every model via stochastic gradient descent (SGD) with momentum of weight $0.9$. The initial learning rate is set to $0.1$, and ``step decay'' is performed during training where the exact scheduling across datasets is specified in the supplementary material. Although it did not affect much to our method, we also adopt the ``linear warm-up'' learning rate strategy \citep{goyal2017accurate} in the first 5 epochs, as the performance of some baseline methods, \eg, re-weighting, highly depends on the use of this strategy. 
For CIFAR-10/100 and CelebA, we train ResNet-32 \citep{he2016identity} for 200 epochs with mini-batch size 128, and set a weight decay of $2 \times 10^{-4}$. In case of SUN397, the pre-activation ResNet-18 model is used instead.\footnote{We remark this model is larger than ResNet-32 used for CIFAR and CelebA datasets, as it has roughly 4$\times$ more channels.}  
We ensure that all the input images are normalized over the training dataset, and have the size of 32$\times$32 either by cropping or re-sizing, to be compatible with the given architectures.
For Twitter and Reuters datasets, we train 2-layer fully-connected networks for 15 epochs with mini-batch size 64, and with a weight decay of $5 \times 10^{-5}$.

\vspace{-0.15in}
\paragraph{Details on M2m.}
When our method is applied, we use another classifier $g$ of the same architecture to $f$ that is pre-trained on the given (imbalanced) dataset via standard ERM training.  
Also, in a similar manner to that of \citet{cao2019learning}, we use the {deferred} scheduling to our method, \ie, we start to apply our method after the standard ERM training for a fixed number of epochs. The actual scheduling across datasets is specified in the supplementary material.
We choose hyperparameters in our method from a fixed set of candidates, namely $\beta \in \{0.9,0.99,0.999\}$, $\lambda \in \{0.01,0.1,0.5\}$ and $\gamma \in \{0.9,0.99\}$ based on the validation set. Unless otherwise stated, we fix $T=10$ and $\eta=0.1$ when performing a single generation step.

\subsection{Long-tailed CIFAR datasets} 

We consider a ``synthetically long-tailed'' variant of CIFAR~\citep{dataset/cifar} datasets (CIFAR-LT-10/100) in order to evaluate our method on various levels of imbalance, where the original datasets are class-balanced.
To simulate the long-tailed distribution frequently appeared in imbalanced datasets, we control the \emph{imbalance ratio} $\rho > 1$ and artificially reduce the training sample sizes of each class except the first class, so that: (a) $N_1/N_K$ equals to $\rho$, and (b) $N_k$ in between $N_1$ and $N_K$ follows an exponential decay across $k$. We keep the test dataset unchanged during this process, \ie, it is still perfectly balanced, thereby measuring accuracy on this dataset is equivalent to measuring the balanced accuracy.
We consider two imbalance ratios $\rho\in\{100, 10\}$ each for CIFAR-LT-10 and 100. See Figure~\ref{fig:CIFAR-10} and \ref{fig:CIFAR-100} for a detailed illustration of the sample distribution. 
 
\begin{table*}[t]
	\vspace{-0.05in}
	\begin{center}
	\begin{tabular}{cccccccccc}
		\toprule
		\multicolumn{2}{c}{Dataset}   &  \multicolumn{4}{c}{CIFAR-LT-10} &  \multicolumn{4}{c}{CIFAR-LT-100} \\
		\cmidrule(r){3-6} \cmidrule(l){7-10}
		\multicolumn{2}{c}{Imbalance ratio}   &  \multicolumn{2}{c}{$N_1/N_K=100$} &  \multicolumn{2}{c}{$N_1/N_K=10$}  &  \multicolumn{2}{c}{$N_1/N_K=100$} &  \multicolumn{2}{c}{$N_1/N_K=10$} \\
        \cmidrule(r){3-4} \cmidrule(l){5-6} \cmidrule(r){7-8} \cmidrule(l){9-10}
		Loss      &   Re-balancing    &   bACC & \echo{GM}    &   bACC  & \echo{GM}  &   bACC & \echo{GM}    &   bACC  & \echo{GM}
		\\ \midrule
		ERM       &       -         &    68.7\ms{1.43} &  \echo{{66.4}\ms{1.69}}     
		          &   86.0\ms{0.69}  &  \echo{85.8\ms{0.50}}  &   37.2\ms{1.12}    &  \echo{21.5\ms{1.66}}     
		          &    56.2\ms{0.69}  &  \echo{51.8\ms{0.63}}  \\ 
        ERM       &   RS            &   70.4\ms{1.15}  &  \echo{69.0\ms{1.36}} 
		          &     86.6\ms{0.37} &  \echo{86.4\ms{0.37}} &  31.6\ms{1.26}   &  \echo{17.7\ms{1.33}}    
		          &  54.8\ms{0.47}  &  \echo{50.3\ms{0.68}} \\
		ERM       &   SMOTE         &   71.5\ms{0.57}  &  \echo{70.2\ms{0.93}} 
		          &    85.7\ms{0.25} &  \echo{85.5\ms{0.26}} &   34.0\ms{0.33}  &  \echo{19.6\ms{0.36}}     
		          &   53.8\ms{0.93}    &  \echo{49.4\ms{1.15}}  \\
		ERM       &   RW           &   72.8\ms{0.33}   &  \echo{72.0\ms{0.29}}
		          &   86.6\ms{0.18}  &  \echo{86.5\ms{0.16}} &   30.1\ms{0.59}   &  \echo{17.6\ms{0.85}}    
		          &   56.0\ms{0.35}    &  \echo{52.0\ms{0.51}}    \\
		ERM       &   CB-RW        &   71.2\ms{1.14} &  \echo{70.0\ms{1.28}}   
		          &    86.8\ms{0.49}  &  \echo{86.6\ms{0.53}} &     38.6\ms{0.46}   &  \echo{22.5\ms{0.49}}    
		          &     55.9\ms{0.24}    &  \echo{52.0\ms{0.42}} \\
		ERM       &   DRS           &   \underline{75.2}\ms{0.26}   &  \echo{\underline{73.9}\ms{0.17}}
		          &  \underline{87.1}\ms{0.26}  &  \echo{\underline{87.0}\ms{0.29}} &   \underline{41.5}\ms{0.21}  &  \echo{\underline{31.0}\ms{0.21}}      
		          &    \underline{57.7}\ms{0.40}   &  \echo{\underline{54.8}\ms{0.33}}  \\
		\textbf{ERM} &   \textbf{M2m (ours)} &   \textbf{78.3}\ms{0.16} &  \echo{\textbf{77.8}\ms{0.16}}
		          &  \textbf{87.9}\ms{0.21}  &  \echo{\textbf{87.5}\ms{0.15}} &  \textbf{42.9}\ms{0.16}  &  \echo{\textbf{33.0}\ms{0.11}}    
		          &    \textbf{58.2}\ms{0.08}   &  \echo{\textbf{55.3}\ms{0.05}}  \\ \midrule
		Focal     &       -         &     68.3\ms{1.19}  &  \echo{{65.5}\ms{1.71}}
		          &  85.3\ms{0.47}  &  \echo{85.1\ms{0.47}} &   37.7\ms{1.38}   &  \echo{22.1\ms{1.49}}    
		          &   55.3\ms{0.42}   &  \echo{\underline{50.7}\ms{0.43}}    \\
		LDAM      &       -         &   72.8\ms{0.37} &  \echo{70.8\ms{0.65}}    
		          &   86.2\ms{0.12}  &  \echo{86.0\ms{0.15}}   &   {39.5}\ms{0.69}  &  \echo{20.8\ms{0.49}}        
		          &   54.7\ms{0.16}   &  \echo{44.1\ms{0.53}}     \\
		LDAM      &   DRW           &  \underline{77.1}\ms{0.49} &  \echo{\underline{76.7}\ms{0.59}}     
		          &   \underline{87.1}\ms{0.28}  &  \echo{\underline{86.9}\ms{0.28}}   &    \underline{42.1}\ms{0.09}      &  \echo{\underline{29.2}\ms{0.27}}   
		          &   \underline{56.9}\ms{0.15}   &  \echo{50.4\ms{0.29}}    \\ 
		\textbf{LDAM} &   \textbf{M2m (ours)}  &  \textbf{79.1}\ms{0.19}  &  \echo{\textbf{78.6}\ms{0.19}}
		          &  \textbf{87.5}\ms{0.15} &  \echo{\textbf{87.4}\ms{0.19}}  &   \textbf{43.5}\ms{0.22}  &  \echo{\textbf{34.2}\ms{0.62}}    
		          &   \textbf{57.6}\ms{0.14}  &  \echo{\textbf{51.8}\ms{0.38}}  
		\\ \bottomrule
	\end{tabular}
    \end{center}
    \vspace{-0.05in}
    \caption{Comparison of classification performance on the four different types of long-tailed CIFAR-10/100 datasets.}
	\label{table:cifar}
    \vspace{-0.02in}
\end{table*}
\begin{table*}[t]
	\begin{center}
	\begin{tabular}{cccccccccccccc}
		\toprule
		\multicolumn{2}{c}{Datasets}   &  \multicolumn{2}{c}{CelebA-5} &  \multicolumn{2}{c}{SUN397} &  \multicolumn{2}{c}{Twitter} &  \multicolumn{2}{c}{Reuters}  \\
		\cmidrule(r){3-4} \cmidrule(l){5-6} \cmidrule(l){7-8} \cmidrule(l){9-10}
		\multicolumn{2}{c}{Imbalance ratio}   & 
		\multicolumn{2}{c}{$N_1/N_K \approx 10.7$} &  
		\multicolumn{2}{c}{$N_1/N_K \approx 46.2$} &  \multicolumn{2}{c}{$N_1/N_K \approx 147.9$} &  \multicolumn{2}{c}{$N_1/N_K = 710$} \\
        \cmidrule(r){3-4} \cmidrule(l){5-6} \cmidrule(l){7-8}  \cmidrule(l){9-10}
		Loss      &   Re-balancing    &   bACC   & \echo{GM} &   bACC   & \echo{GM}   &   bACC  & \echo{GM} &   bACC  & \echo{GM} \\ \midrule
		ERM       &       -         
		          &    72.7\ms{1.24}  &  \echo{69.4\ms{0.97}}
		          &    31.5\ms{0.07}  &  \echo{20.2\ms{0.74}}
		          &    74.7\ms{0.46}  &  \echo{65.2\ms{1.10}}  
		          &    59.8\ms{1.17}  & \echo{53.8\ms{1.75}}  \\ 
        ERM       &   RS  
                  &    72.5\ms{0.93}   & \echo{70.4\ms{1.37}}
                  &    28.4\ms{0.19}   & \echo{19.8\ms{1.10}}
                  &    75.8\ms{0.30}   & \echo{70.4\ms{1.67}}
		          &    63.3\ms{0.90} &  \echo{57.4\ms{1.03}} \\
		ERM       &   SMOTE         
		          &    72.8\ms{1.07}  & \echo{70.7\ms{0.84}}
		          &    23.7\ms{0.09}  & \echo{14.8\ms{0.39}}
		          &    75.8\ms{0.38}  & \echo{69.5\ms{0.30}}
		          &    62.5\ms{1.30}  & \echo{56.8\ms{1.69}} \\
		ERM       &   RW           
		          &    \underline{74.5}\ms{0.50}  &  \echo{\underline{73.4}\ms{0.87}}
		          &    31.3\ms{0.20}  &  \echo{\underline{25.3}\ms{0.12}}
		          &    76.2\ms{0.95}  &  \echo{73.5\ms{1.46}}
		          &    \underline{65.0}\ms{1.08}  &  \echo{\underline{59.2}\ms{1.84}}\\
		ERM       &   CB-RW           
		          &   74.2\ms{0.59}  & \echo{72.3\ms{0.50}}
		          &   \underline{31.7}\ms{0.13}  & \echo{25.1\ms{0.51}}
		          &   77.5\ms{0.40}  & \echo{73.6\ms{0.79}}
		          &    64.8\ms{0.45} & \echo{57.6\ms{1.62}}  \\
		ERM       &   DRS           
		          &   73.1\ms{0.68}  & \echo{71.2\ms{0.62}} 
		          &   30.7\ms{0.34}  & \echo{24.2\ms{0.40}} 
		          &     \underline{77.8}\ms{0.85} &  \echo{\underline{74.3}\ms{1.48}}
		          &    62.4\ms{0.39} & \echo{56.0\ms{1.34}}  \\
		\textbf{ERM} &   \textbf{M2m (ours)} 
		          &   \textbf{75.6}\ms{0.16} & \echo{\textbf{74.6}\ms{0.34}}
		          &   \textbf{32.4}\ms{0.17} & \echo{\textbf{25.8}\ms{0.29}}
		          &   \textbf{78.2}\ms{0.35} & \echo{\textbf{74.8}\ms{0.78}}
		          &   \textbf{66.3}\ms{0.42}  & \echo{\textbf{60.5}\ms{0.52}} \\ \midrule
		Focal     &       -         
		          &  72.7\ms{0.57} &  \echo{69.7\ms{1.42}} 
		          &  31.2\ms{0.14} &  \echo{21.3\ms{0.71}} 
		          &  74.2\ms{2.35} &  \echo{70.4\ms{4.03}} 
		          &     59.4\ms{0.42}  & \echo{53.0\ms{0.74}}  \\
		LDAM      &       -         
		          &   73.0\ms{1.14} &  \echo{68.0\ms{2.19}}  
		          &   30.2\ms{0.10} &  \echo{14.4\ms{0.83}}   
		          &   74.6\ms{0.40} &  \echo{66.1\ms{2.28}}   
		          &    63.0\ms{1.36}  &  \echo{\underline{57.6}\ms{0.50}} \\ 
		LDAM      &   DRW           
		          &   \underline{74.4}\ms{0.33} &  \echo{\underline{72.3}\ms{0.82}}   
		          &   \underline{31.6}\ms{0.10} &  \echo{\underline{23.6}\ms{0.36}}   
		          &   \underline{78.0}\ms{0.87}  &  \echo{\underline{74.4}\ms{1.28}}
		          &   \underline{64.1}\ms{0.31} & \echo{56.9\ms{1.08}}   \\
		\textbf{LDAM} &   \textbf{M2m (ours)}  
		          &   \textbf{75.9}\ms{1.09}  & \echo{\textbf{75.0}\ms{0.94}}
		          &   \textbf{33.3}\ms{0.20}  & \echo{\textbf{24.9}\ms{0.76}}
		          &   \textbf{78.8}\ms{0.21}  & \echo{\textbf{76.0}\ms{0.23}}
		          &     \textbf{70.0}\ms{0.68}    & \echo{\textbf{63.9}\ms{0.49}}
		\\ \bottomrule
	\end{tabular}
    \end{center}
    \vspace{-0.05in}
    \caption{Comparison of classification performance on the four naturally imbalanced datasets: CelebA-5, SUN397, Twitter and Reuters. In case of Reuters, $\eta$ is adjusted to 1.0 when training M2m models regarding the numerical range of the dataset.}
	\label{table:nlp_tasks}
    \vspace{-0.1in}
\end{table*}

Table \ref{table:cifar} summarizes the main results.
In overall, the results show that our method consistently improves the bACC by a large margin, across all the tested baselines. 
These results even surpass the ``LDAM+DRW'' baseline \citep{cao2019learning}, which is known to be the state-of-the-art to the best of our knowledge. 
Moreover, we point out, in most cases, our method could further improve bACC when applied upon the LDAM training scheme (see ``LDAM+M2m''): this indicates that the performance gain from our method is fairly orthogonal to that of LDAM, \ie, the margin-based approach, which suggests a new promising direction of improving the generalization when a neural network model suffers from a problem of small data. 

\vspace{-0.02in}
\subsection{Real-world imbalanced datasets}
\label{ss:real}

We further verify the effectiveness of M2m on four well-known, \emph{naturally} imbalanced datasets: 
CelebA \citep{liu2015faceattributes}, SUN397 \citep{xiao2010sun}, Twitter \citep{gimpel2010part} and Reuters \citep{lewis2004rcv1} datasets. More detailed information for each of these datasets is demonstrated in Figure~\ref{fig:stat} and the supplementary material.

CelebA is originally a multi-labeled dataset, and we port this to a 5-way classification task by filtering only the samples with five non-overlapping labels about hair colors. We also subsampled the full dataset by $1/20$ while maintaining the imbalance ratio $\rho\approx 10.7$, in attempt to make the task more difficult. We denote the resulting dataset by CelebA-5.

Although Twitter and Reuters datasets are from natural language processing, we also evaluate our method on them to test the effectiveness under much extreme imbalance. Here, we remark that the imbalance ratio $N_{1}/N_{k}$ of these two datasets are about 150 and 710, respectively, which are much higher than the other image datasets tested. In case of Reuters, we exclude the classes having less than 5 samples in the test set for more reliable evaluation, resulting a dataset of 36 classes. 

Table~\ref{table:nlp_tasks} shows the results. Again, M2m performs best amongst other baseline methods, demonstrating the effectiveness of our method under natural imbalance, as well as {wider} applicability of our algorithm beyond image classification. Remarkably, the significant results on Reuters dataset compared to the others suggest that our method can be even more effective under a regime of ``extremely'' imbalanced datasets, as Reuters has a much larger imbalance ratio than the others.

\subsection{Ablation study}
\label{ss:ablation}

We conduct an extensive ablation study to present a detailed analysis of the proposed method. All the experiments in this section are performed with ResNet-32 models, trained on CIFAR-LT-10 with the imbalance ratio $\rho=100$. We additionally report the balanced test accuracy over \emph{majority} and \emph{minority} classes, namely ``Major'' and ``Minor'' respectively, to further identify the relative impacts on those two classes separately. We divide the whole classes into ``majority'' and ``minority'' classes, so that the majority classes consist of top-$k$ frequent classes with respect to the training set where $k$ is the minimum number that $\sum_k N_k$ exceeds $50\%$ of the total. We denote the minority classes as the remaining classes. We provide more discussion in the supplementary material.

\begin{table}[t]
    \begin{center}
    \small
	\begin{tabular}{ccc}
		\toprule
		\# Seeds             &  bACC ($\Delta$)  &  {GM ($\Delta$)} \\ \midrule
		10           &   {74.9}{\ms{0.29}} {(-4.34\%)}  & {73.7{\ms{0.33}} {(-5.27\%)}}   \\
		50           &   76.2{\ms{0.30}} (-2.68\%) & {75.3{\ms{0.29}} (-3.21\%)} \\ 
        100          &   76.5{\ms{0.34}} (-2.30\%) & {75.6{\ms{0.41}} (-2.83\%)} \\ 
        200          &   76.7{\ms{0.51}} (-2.04\%) & {75.9{\ms{0.59}} (-2.44\%)} \\ 
        500          &   77.4{\ms{0.38}} (-1.15\%) & {76.8{\ms{0.31}} (-1.29\%)}
        \\ \midrule
        \textbf{Full} &  
        \textbf{78.3\ms{0.16}}  (-0.00\%)  & 
        {\textbf{77.8}\ms{0.16}} (-0.00\%) \\ 
        
        \bottomrule
	\end{tabular}
	\end{center}
	\vspace{-0.05in}
	\caption{Comparison of classification performance across various number of samples allowed to be a seed sample $x_0$. $\Delta$ indicates the relative gap from the original result presented in ``Full''.}
	\label{table:abl2}
	\vspace{-0.05in}
\end{table}

\begin{table}[t]
	\begin{center}
    \begin{adjustbox}{width=1\linewidth}
	\begin{tabular}{lcccc}
		\toprule
		Methods             &   Major (2)      &    Minor (8)    &  bACC  &  {GM} \\ \midrule
        M2m ($\lambda=0$)   &   92.8\ms{0.97}       &    73.0\ms{0.10}      &  76.9\ms{0.15} & {76.5\ms{0.11}} \\ 
        M2m-Clean    &   78.4\ms{2.45}       &    72.7\ms{0.60}      &  73.5\ms{0.81}  & {73.0\ms{0.93}} \\ \midrule
        ERM-RS  &  92.8\ms{1.50}       &    64.8\ms{1.18}      &  70.4\ms{1.15}  &  {69.0\ms{1.36}} \\
        M2m-RS   &   92.9\ms{2.99}       &    69.4\ms{0.84}      &  74.1\ms{0.10}  &   {73.1\ms{0.14}} \\ 
        M2m-RS-Rand   &   93.6\ms{2.34}       &    66.1\ms{1.04}      &  71.6\ms{0.36}  &   {70.3\ms{0.80}} \\ \midrule
        \textbf{M2m}         &   \textbf{93.3}\ms{0.85}       &    \textbf{74.6}\ms{0.34}      &  \textbf{78.3}\ms{0.16} & {\textbf{77.8}\ms{0.16}}    \\ 
        \bottomrule
	\end{tabular}
	\end{adjustbox}
	\end{center}
	\vspace{-0.05in}
	\caption{Comparison of classification performance across various types of ablations. We report the number of majority and minority classes in the parentheses.}
	\label{table:abl1}
	\vspace{-0.15in}
\end{table}

\vspace{-0.15in}
\paragraph{Diversity on seed samples.} In Section~\ref{ss:overview}, we hypothesize that the effectiveness of our method mainly comes from utilizing a much diversity in the majority samples to prevent the over-fitting to the minority classes. To verify this, we consider an ablation that the candidates of ``seed samples'' are limited: more concretely, we control the size of seed sample pools per each class to a fixed subset of the training set, made before training $f$. In Table~\ref{table:abl2}, the accuracy of minority classes is progressively increased as seed sample pools become diverse. This clear trend indicates that M2m makes use of the diversity of majority classes for preventing the over-fitting to the minority classes. 

\vspace{-0.15in}
\paragraph{The effect of $\lambda$.} In the optimization objective \eqr{eq:obj} for the generation step in M2m, we impose a regularization term $\lambda\cdot f_{k_0}(x)$ to improve the quality of synthetic samples: they might confuse $f$ if themselves still contain important features of the original class in a viewpoint of $f$. To verify the effect of this term, we consider an ablation that $\lambda$ is set to $0$, and compare the performance to the original method. As reported in Table~\ref{table:abl1}, we found a certain level of degradation in the balanced test accuracy at this ablation, which shows the effectiveness of the proposed regularization.

\vspace{-0.15in}
\paragraph{Over-sampling from the scratch.} 
As specified in Section~\ref{ss:setup}, we use the ``deferred'' scheduling to our method by default, \ie, we start to apply our method after the standard ERM training for a fixed number of epochs. We have also considered a simple ablation where this strategy is not used, namely ``M2m-RS''. The results in Table~\ref{table:abl1} show that M2m-RS still outperforms any other baselines (reported in Table~\ref{table:cifar}) except the ones that the deferred scheduling is used, \ie, DRS and DRW, and this further verifies the effectiveness of our method.

\vspace{-0.15in}
\paragraph{Labeling as a targeted class.} 
Our primary assumption on the pre-trained classifier $g$ does not require that $g$ itself to generalize well on the minority classes (see Section~\ref{ss:overview}). This implies that solving \eqr{eq:obj} with $g$ may not end up with a synthetic sample that contains generalizable features of the target minority class. To examine how much the generated samples would be correlated to the target classes, we consider another ablation upon M2m-RS:\footnote{Here, we attempt to opt out any potential effect from using DRS, for more clearer evaluation.} 
instead of labeling the generated sample as the target class, the ablated method ``M2m-RS-Rand'' labels it to a ``random'' class chosen from all the possible classes (except for the target and original classes). The results shown in Table~\ref{table:abl1} indicate that M2m-RS-Rand generalizes much worse than its counterpart M2m-RS on the minority classes, which indeed confirms that the correctly-labeled synthetic samples could improve the generalization of the minority classes. 

\begin{figure*}[t]
\begin{center}
    {
    \subfigure[ERM]
        {
        \includegraphics[width=0.19\textwidth]{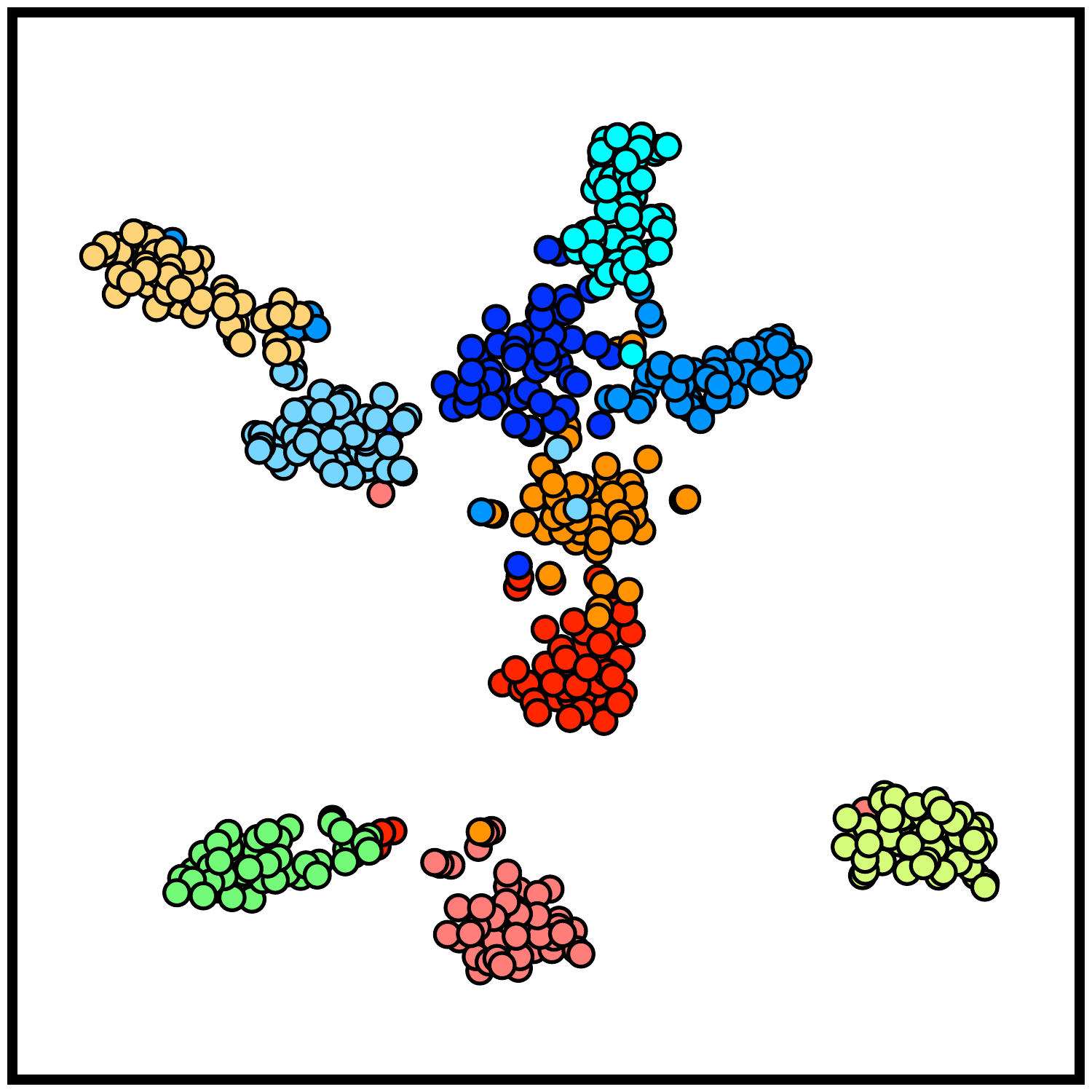}
        \label{fig:tSNE-ERM}}
        \subfigure[ERM-SMOTE]
        {
        \includegraphics[width=0.19\textwidth]{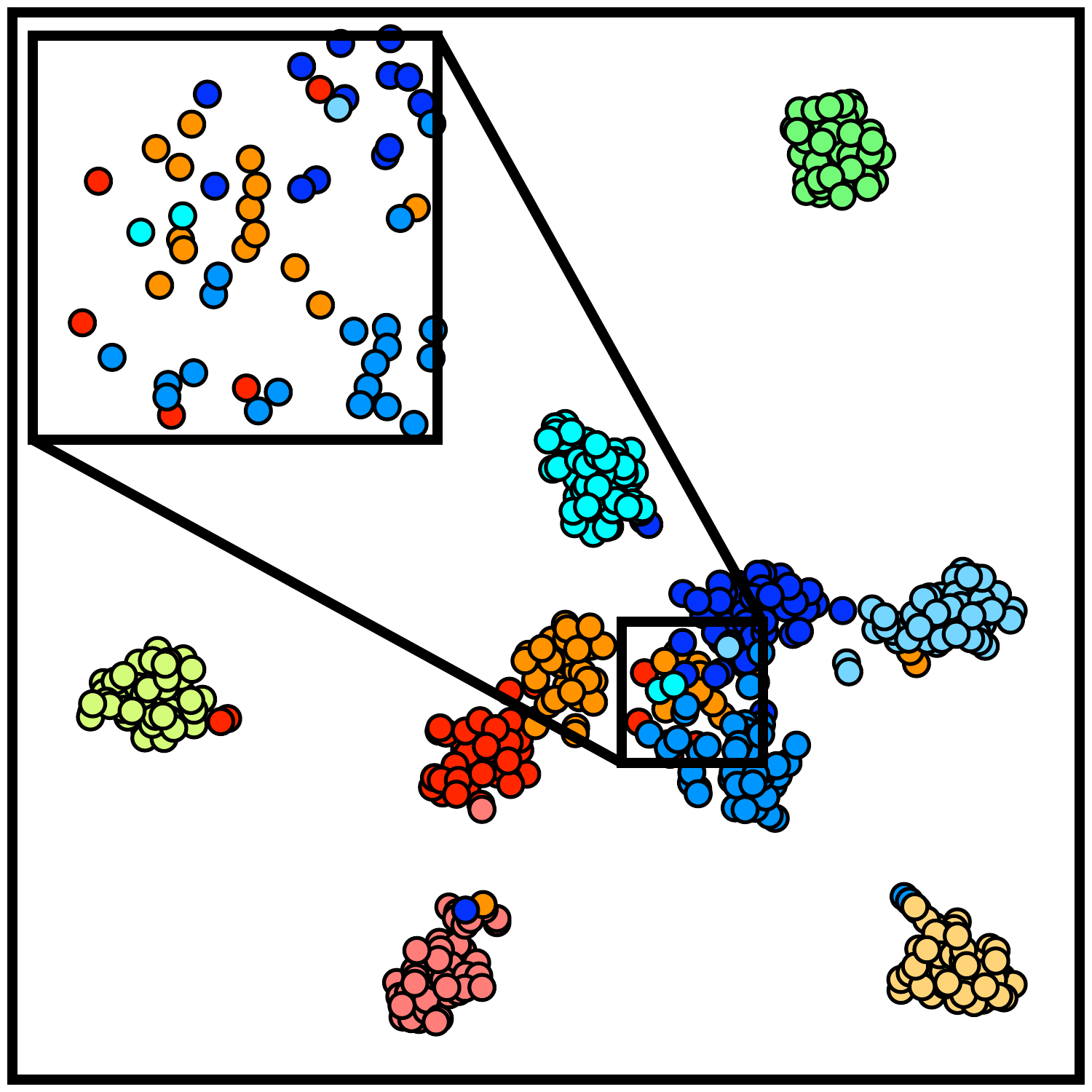}
        \label{fig:tSNE-SMOTE}} 
        \subfigure[LDAM-DRW]
        {
        \includegraphics[width=0.19\textwidth]{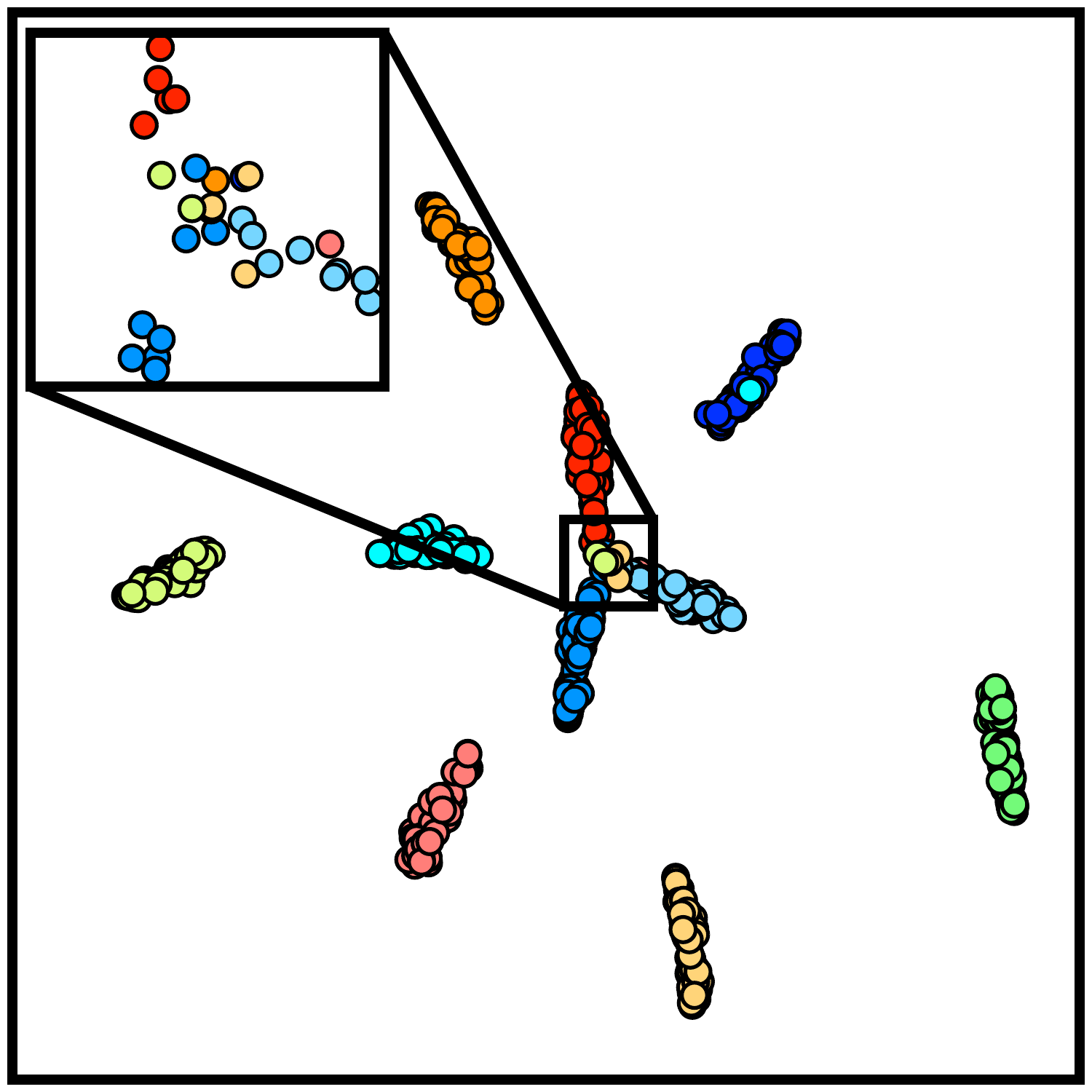}
        \label{fig:tSNE-LDAM}}
        \subfigure[ERM-M2m (ours)]
        {
        \includegraphics[width=0.19\textwidth]{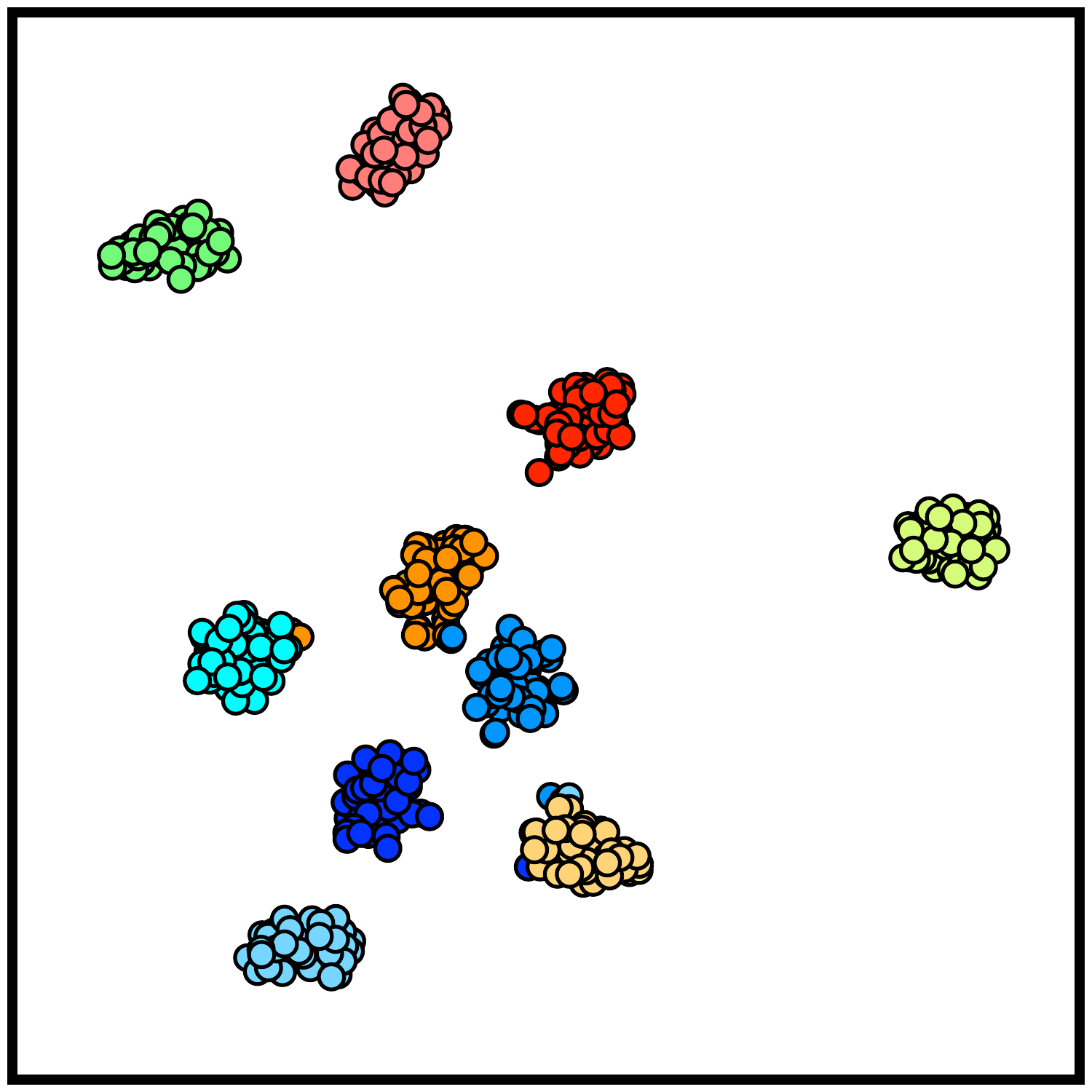}
        \label{fig:tSNE-AO}
        }
    }
\end{center}
\vspace{-0.12in}
\caption{Visualization of the penultimate features via t-SNE computed from a balanced subset of CIFAR-LT-10 with ResNet-32.}
\label{fig:tSNE}
\vspace{-0.12in}
\end{figure*}

\vspace{-0.15in}
\paragraph{Comparison of t-SNE embeddings.} 
To further validate the effectiveness of our method, we visualize and compare the penultimate features learned from various training methods (including ours) using t-SNE \citep{maaten2008visualizing}. Each embedding is computed from a randomly-chosen subset of training samples in the CIFAR-LT-10 ($\rho=100$), so that it consists of 50 samples per each class.
Figure~\ref{fig:tSNE} illustrates the results, and shows that the embedding from our training method (M2m) is of much separable features compared to other methods:
one could successfully distinguish each cluster under the M2m embedding (even though they are from minority classes), while others have some obscure region.

\begin{figure}[t]
    \begin{center}
    \subfigure[CIFAR-LT-10]
        {
        \includegraphics[width=0.47\linewidth]{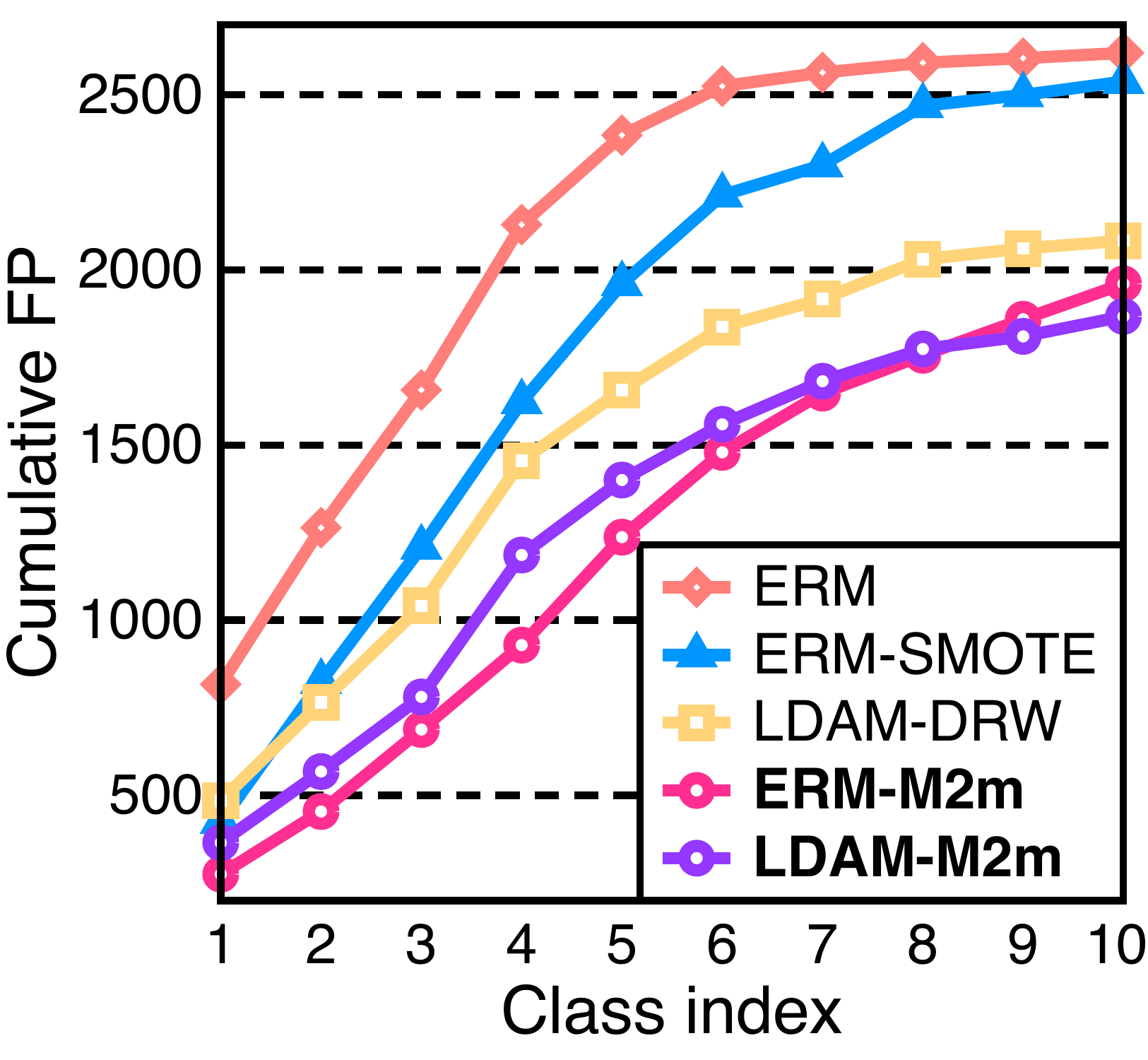}
        \label{fig:cumul-10}
        }
    \subfigure[CIFAR-LT-100]
        {
        \includegraphics[width=0.47\linewidth]{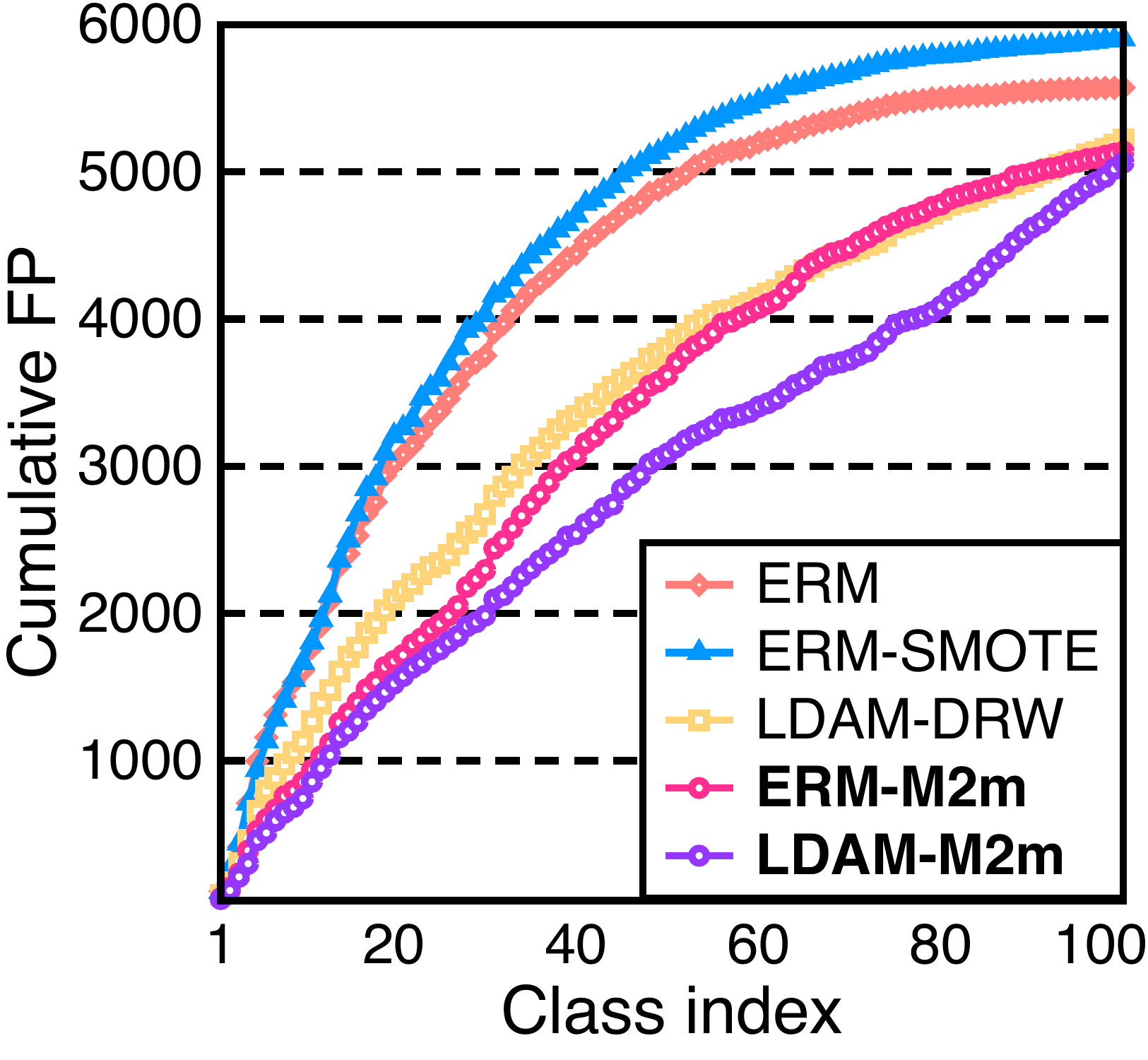}
        \label{fig:cumul-100}
        }
    \end{center}
    \vspace{-0.05in}
    \caption{Comparisons on cumulative number of false positive samples across class indices ($\sum_k \mathrm{FP}_k$) on CIFAR-LT-10/100 test set in ResNet-32. Each plot reaches the total number of mistakes, \ie, the sum of off-diagonal entries in the confusion matrix.}
    \label{fig:cumul}
    \vspace{-0.15in}
\end{figure}

\vspace{-0.15in}
\paragraph{Comparison of cumulative false positive.} In Figure ~\ref{fig:cumul}, we plot how the number of false positive (FP) samples increases as summed over classes, namely  $\sum_k \mathrm{FP}_k$, from the most frequent class to the least one. Here, $\mathrm{FP}_k$ indicates the number of misclassified samples by predicting them to class~$k$ in the test set. We compute each plot with the \emph{balanced} test sets of CIFAR-LT-10/100, thereby a well-trained classifier would show a plot close to linear: it indicates the classifier mistakes more evenly over the classes. Overall, one could see that the curve made by our method consistently below the others with much linearity. This implies our method makes less false positives, and even better, they are more uniformly distributed over the classes. This is a desirable property in the context of imbalanced learning.

\begin{figure}[t]
    \begin{center}
	    \includegraphics[width=0.75\linewidth]{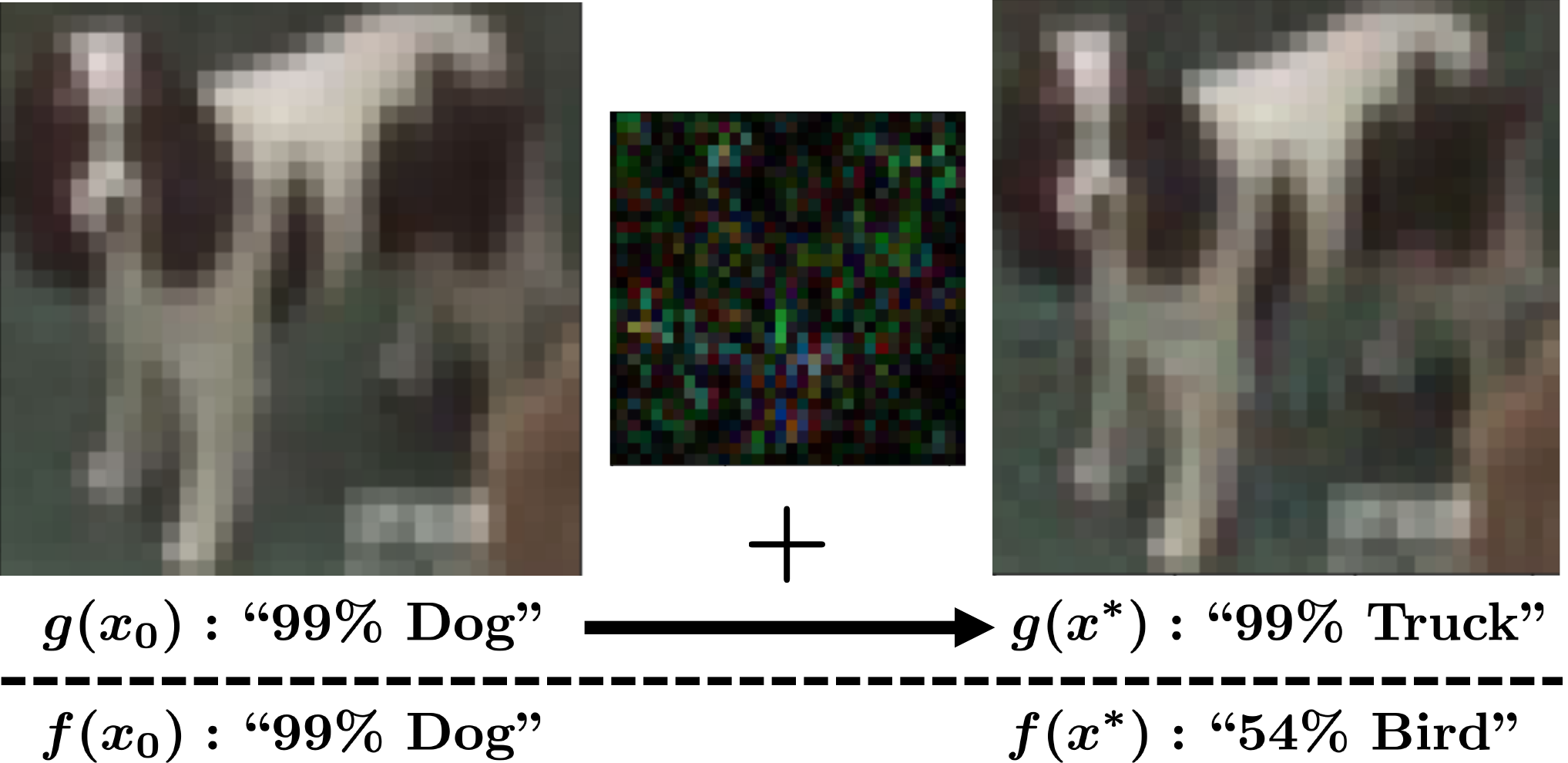}
    \end{center}
    \vspace{-0.12in}
    \caption{An illustration of a synthetic minority sample by M2m, where $g$ is assumed to be ResNet-32 trained by standard ERM. The noise image is amplified by 10 for better visibility.}
    \label{fig:generation}
    \vspace{-0.15in}
\end{figure}

\vspace{-0.15in}
\paragraph{The use of adversarial examples.}
As mentioned in Section~\ref{ss:intuition}, the generation under M2m often ends up with a synthetic minority sample that is very close to the original (before translation) as like the \emph{adversarial example}. This indeed happens when $f$ and $g$ are neural networks as assumed here, \ie, ResNet-32, as illustrated in Figure~\ref{fig:generation}.
To understand more on how such adversarial perturbations affect our method, we consider a simple ablation, which we call ``M2m-Clean'': recall that our method synthesizes a minority sample $x^*$ from a seed majority sample $x_0$. This ablation uses the ``clean'' $x_0$ instead of $x^*$ for over-sampling.
Under the identical training setup, we notice a significant reduction in the balanced accuracy of M2m-Clean compared to the original M2m (see Table~\ref{table:abl1}). This observation reveals that the adversarial perturbations ablated are extremely crucial to make our method to work, regardless of a small noise. 

\vspace{-0.04in}
\section{Conclusion}
\vspace{-0.04in}

We propose a new over-sampling method for imbalanced classification, called \emph{Major-to-minor Translation} (M2m). 
We found the diversity in majority samples could much help the class-imbalanced training, even with a simple translation method using a pre-trained classifier. This suggests a promising way to overcome the long-standing class-imbalance problem, and exploring more powerful methods to perform this Major-to-minor translation, \eg, CycleGAN \cite{zhu2017unpaired}, would be an interesting future research.
The problems we explored in this paper also lead us to an essential question that whether an adversarial perturbation could be a good feature.
Our findings suggest that it could be, at least for the purpose of imbalanced learning, where the minority classes suffer over-fitting due to insufficient data. 
We believe our method could open up a new direction of research both in imbalanced learning and adversarial examples. 

\vspace{-0.04in}
\section*{Acknowledgements}
\vspace{-0.04in}
This work was supported by Samsung Electronics and Institute of Information \& communications Technology Planning \& Evaluation (IITP) grant funded by the Korea government (MSIT) (No.2019-0-00075, Artificial Intelligence Graduate School Program (KAIST)).

{\small
\bibliographystyle{ieee_fullname}
\bibliography{references}
}

\onecolumn
\clearpage
\appendix

\begin{center}{\bf {\LARGE Supplementary Material}}
\end{center}
\begin{center}{\bf {\Large \emph{M2m}: Imbalanced Classification via Major-to-minor Translation}}
\end{center}

\vspace{0.05in}

\section{Details on the datasets}

\paragraph{CIFAR-LT-10/100.} 

CIFAR-10/100 datasets \citep{dataset/cifar} consist of 60,000 RGB images of size $32\times32$, 50,000 for training and 10,000 for testing. Each image in the two datasets is corresponded to one of 10 and 100 classes, respectively. In our experiments, we construct ``{synthetically long-tailed}'' variants of CIFAR-10/100, namely CIFAR-LT-10/100, respectively \citep{cao2019learning}. We hold-out 10\% of the test set to construct a validation set, and use the remaining for testing. We use ResNet-32 \cite{he2016identity} with a mini-batch size 128, and set a weight decay of $2\times10^{-4}$. We train the network for 200 epochs with an initial learning rate of 0.1. We follow the learning rate schedule used by \citet{cui2019class} for fair comparison: the initial learning rate is set to $0.1$, and we decay it by a factor of $100$ at $160$-th and $180$-th epoch. When the deferred scheduling \citep{cao2019learning} is used, \eg, DRS, DRW and our method, it is applied after 160 epochs of standard training.

\vspace{-0.1in}
\paragraph{CelebA-5.}

CelebFaces Attributes (CelebA) dataset \citep{liu2015faceattributes} is a multi-labeled face attributes dataset. It is originally composed of 202,599 number of RGB face images with 40 binary attributes annotations per image. We port this CelebA to a 5-way classification task by filtering only the samples with five non-overlapping labels about hair colors: namely, ``blonde'', ``black'', ``bald'', ``brown'', and ``gray''. This is in a similar manner as done in \cite{Mullick_2019_ICCV}. We denote the resulting dataset by CelebA-5. We pick out 50 and 100 samples per each class for validation and testing. We use ResNet-32 \cite{he2016identity} with a mini-batch size 128, and set a weight decay of $2\times10^{-4}$. We train the network for 90 epochs with an initial learning rate of 0.1. We decay the learning rate by 0.1 at epoch 30 and 60. 
When the deferred scheduling is used, it is applied after 60 epochs of standard training.

\vspace{-0.1in}
\paragraph{SUN397.}

Scene UNderstanding (SUN) \citep{xiao2010sun} is a dataset for a scene categorization. It originally consists of 108,754 RGB images which are labeled with 397 classes. For the inputs, center patches are first extracted and they are resized to 32$\times$32. We hold-out 10 and 40 samples per each class for validation and testing, respectively, as the dataset itself does not provide any separated split for testing. We use pre-activation ResNet-18 \cite{he2016identity} which roughly has $4\times$ more channels with a mini-batch size 128, and set a weight decay of $2\times10^{-4}$. We train the network for 90 epochs with an initial learning rate of 0.1. We decay the learning rate by 0.1 at epoch 30 and 60. When the deferred scheduling is used, it is applied after 60 epochs of standard training.

\vspace{-0.1in}
\paragraph{Twitter.}

Twitter \citep{gimpel2010part} is a dataset for a part-of-speech (POS) tagging task in social media text with 25 classes. Each sample is a pair of a token and a tag, \eg, ``(books, common noun)'' and ``(\#acl, hashtag)'', where each token is embedded into a 50-dimensional vector via a pre-defined word-embedding \citep{hendrycks2016baseline}. We discarded two classes with zero test samples and obtained 14,614 training samples with 23 classes. We use 2-layer fully-connected network with a hidden layer size of 256 and a ReLU nonlinearity. We set a mini-batch size 64 and a weight decay of $5\times10^{-5}$. We train the network for 15 epochs with an initial learning rate 0.1 and decay the learning rate by 0.1 at epoch 10. When the deferred scheduling is used, it is applied after 10 epochs of standard training.  

\vspace{-0.1in}
\paragraph{Reuters.}

Reuters \citep{lewis2004rcv1} is a dataset for a text categorization task which predicts the subject of a given text. As an input, 1000-dimensional bag-of-words vectors are given, which are processed from a news story document. It is originally composed of 52 classes, but we discarded the classes that have less than 5 test samples for a reliable evaluation, obtaining a subset of the full dataset of 36 classes with 6436 training samples. We hold-out 10\% of training samples to construct a validation set. We use 2-layer fully-connected network with a hidden layer size of 256 and a ReLU nonlinearity. We set a mini-batch size 64 and a weight decay of $5\times10^{-5}$. We train the network for 15 epochs with an initial learning rate 0.1 and decay the learning rate by 0.1 at epoch 10. When the deferred scheduling is used, it is applied after 10 epochs of standard training.   

\section{More results from ablation study}
\paragraph{Generation from another classifier $g$.} 
As mentioned, our method introduces another classifier $g$ to generate synthetic minority $x^{*}$ independently from the training classifier $f$.
This is because using $f$ itself instead of $g$ in the optimization objective (2) would let the synthetic samples already confident in the target minority class to $f$, and this makes the overall training process redundant.
To further validate the importance of using $g$, we consider an ablation called ``M2m-Self'': instead of using $g$, ``M2m-Self'' uses $f$ for generating minority samples. 
As reported in Table~\ref{table:supp1}, one could immediately see that M2m-Self only shows 
marginal improvement from DRS, which is much inferior than the original M2m.

\paragraph{Using multiple classifiers for generation.}
 
Since our method is not restricted to use the only one pre-trained classifier $g$ in the optimization (2), the multiple classifiers $g_{i}$ for $i=1,\dots,m$ can be used to improve the quality of generation. To verify the additional gain from multiple classifiers, we consider an ablation called ``M2m-Ensemble'': use the ensemble of the classifiers ($m=2$) for generation instead of the single classifier. Here, we use the same architecture ResNet-32 for $g_{1}$ and $g_{2}$ and use a higher $\gamma$ due to the smoothed prediction from the ensemble. The results in Table~\ref{table:supp1} show that M2m-Ensemble slightly perform better than M2m. It indicates that our method can benefit from the stronger classifier.  

\begin{table}[t]
\begin{minipage}[t]{0.97\textwidth}
    \hfill
    \begin{minipage}[b]{0.51\linewidth}
        \begin{center}
    	\begin{tabular}{lcc}
    		\toprule
    		Methods             &  bACC ($\Delta$)  &  {GM ($\Delta$)} \\ \midrule
    		ERM-DRS           &   {75.2}{\ms{0.26}} {(-3.96\%)}  & {73.9{\ms{0.32}} {(-5.01\%)}} \\
    		M2m-Self           &   {75.9}{\ms{0.27}} {(-3.07\%)}  & {74.9{\ms{0.32}} {(-3.73\%)}}   \\
    		M2m-No-Reject           &   {77.4}{\ms{0.33}} {(-1.15\%)}  & {76.8{\ms{0.40}} {(-1.29\%)}}   \\ 
    		M2m ($\gamma = 0$)           &   76.9{\ms{0.19}} (-1.79\%) & {76.4{\ms{0.20}} (-1.80\%)} \\ \midrule
            {M2m} &  
            {78.3\ms{0.16}} (-0.00\%)  & 
            {77.8\ms{0.16}} (-0.00\%) \\
            M2m-Ensemble & {78.5\ms{0.20}}  (+0.26\%)  & {{78.0}\ms{0.22}} (+0.26\%) \\
            \bottomrule
    	\end{tabular}
    	\end{center}
    	\caption{Comparison of classification performance across various types of ablations. $\Delta$ indicates the relative gap from the original result presented in ``M2m''. All the values and error bars are mean and standard deviation across three random trials, respectively.}
    	\label{table:supp1}
    \end{minipage}
    \hspace{0.25in}
    \begin{minipage}[b]{0.42\linewidth}
        \begin{center}
        \begin{tabular}{cccc}
        \toprule
        Loss    & Re-balancing         &  bACC ($\Delta$)  &  {GM ($\Delta$)}  \\ \midrule
        ERM           & - &   {38.6}{\ms{0.75}}  & 26.9{\ms{0.78}} \\
        ERM       & DRS    &   {40.8}{\ms{0.67}} & {31.6}{\ms{1.05}}   \\
        \textbf{ERM} & \textbf{M2m (ours)}  &   \textbf{42.2}{\ms{0.51}} & \textbf{33.1}{\ms{0.64}} \\ \midrule
        LDAM & -        &   41.0{\ms{0.07}} & {28.5{\ms{0.11}}} \\ 
        LDAM & DRW        &   43.0{\ms{0.17}} & {34.5{\ms{0.17}}}  \\ 
        \textbf{LDAM} & \textbf{M2m (ours)}  &  \textbf{43.7}\ms{0.26} & {\textbf{35.1}\ms{0.35}}  \\ \bottomrule
        \end{tabular}
        \end{center}
        \caption{Comparison of classification performance on ImageNet-LT. All the values and error bars are mean and standard deviation across three random trials, respectively.}
        \label{table:supple_imagenet}
    \end{minipage}
    \hfill
\end{minipage}
\end{table}

\paragraph{Rejection criteria.} 
We also propose a sample rejection criteria to alleviate the risk of unreliable generation, possibly due to a weak generalization of $g$. To verify the effect of this rejection criteria, we consider an ablation, namely ``M2m-No-Reject'', which does not use this rejection policy in training. In other words, all the generated samples are used to train $f$. The results in Table~\ref{table:supp1} show that M2m-No-Reject performs significantly worse than M2m. This indeed confirms the gain from using the proposed rejection criteria.   

\paragraph{The effect of $\gamma$.} 

As specified in Algorithm~1 in the main paper, we set a threshold $\gamma$ to filter out the synthetic samples which the generation objective is not sufficiently minimized, mainly due to the limited budget. To evaluate the practical effectiveness of using $\gamma$, here we consider an ablation that this thresholding is not used, equivalently when $\gamma=\infty$. As reported in Table~\ref{table:supp1}, we indeed observe a performance degradation by not using $\gamma$. This reveals that the confidence level in $g$ affects the final quality of the generation.

\section{Results on ImageNet-LT}

\begin{wrapfigure}[13]{r}{0.2\textwidth}
	\vspace{-0.8cm}
	\begin{center}{
	\hspace{-0.4cm}
	\includegraphics[width=36mm]{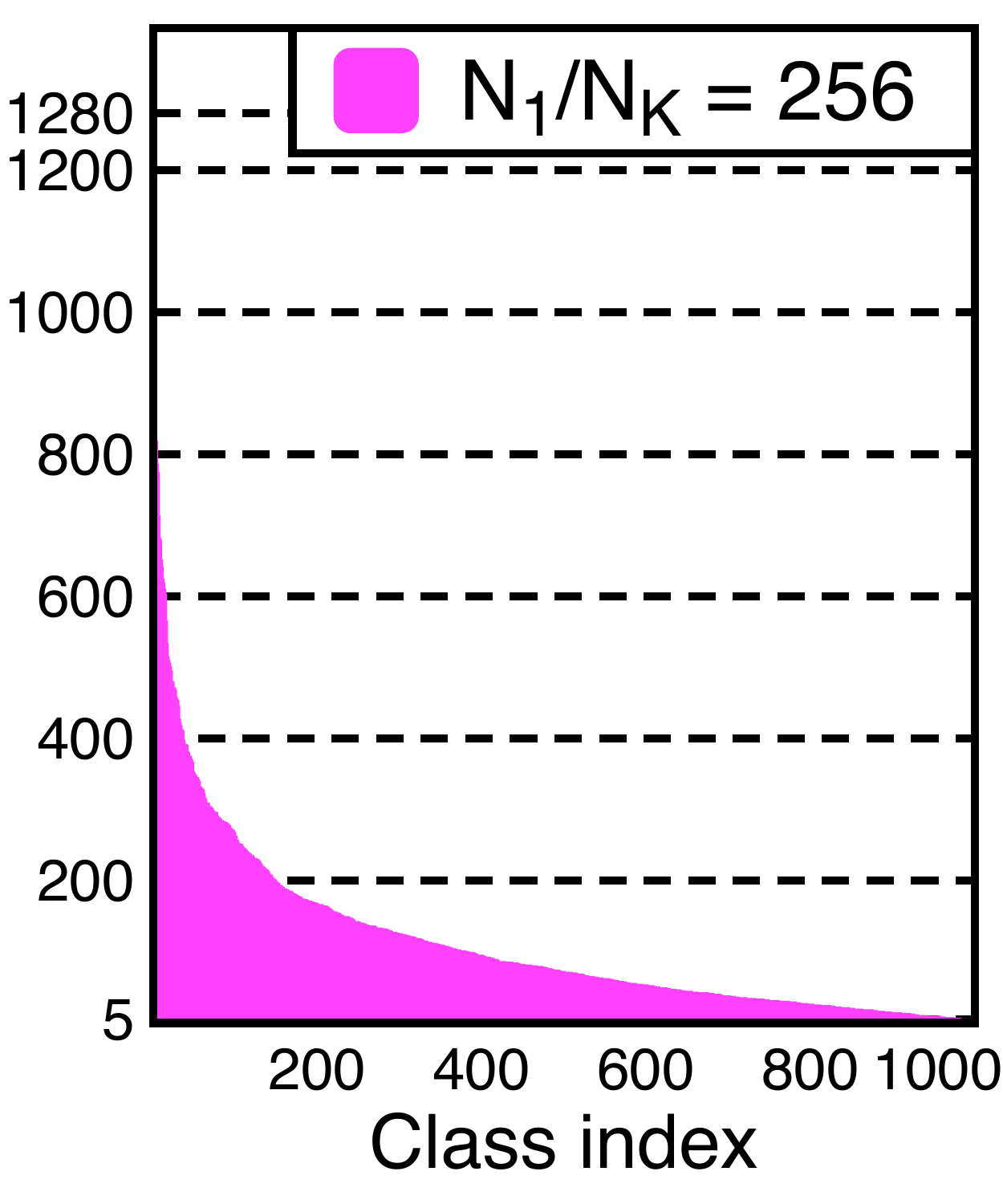}
	\caption{Class distribution of ImageNet-LT.}
	\label{fig:imagenet}
	}\end{center}
\end{wrapfigure}
We additionally evaluate our method on ImageNet-LT \citep{liu2019large} dataset, a subset of ImageNet dataset \cite{ILSVRC15} with a synthetic imbalance following the Pareto distribution of the power $\alpha=6$. It is composed of 115,846 training samples with 1,000 categories, 1,280 images in the maximal class and 5 images in the minimal class. A more detailed distribution is presented in Figure~\ref{fig:imagenet}. We use the randomly-resized cropping and the horizontal flipping as a data augmentation, and all the images are resized to 128$\times$128. We hold-out 20 samples per class randomly from the original ImageNet training set to form a validation set, and the original (roughly balanced) ImageNet validation set is used for testing. We use ResNet-50 \cite{he2016deep} with a mini-batch size 256 and set a weight decay of $10^{-4}$. We train the network for 200 epochs with an initial learning rate of 0.1 and it is decayed by 0.1 at epoch 160 and 180. When the deferred scheduling is used, \eg, DRS, DRW and our method, it is applied after 160 epochs of standard training. We evaluate our method with followings which show the best performance among the baselines in the experiments in the main paper: (\textit{a}) ERM-DRS and (\textit{b}) LDAM-DRW \citep{cao2019learning}. We report the \emph{balanced accuracy} (bACC) and the \emph{geometric mean scores} (GM). As reported in Table~\ref{table:supple_imagenet}, our method, M2m, significantly outperforms the baselines. In the case of ERM loss, compare to DRS, M2m shows 3.43 \% and 4.75 \% relative gains in bACC and GM, respectively. Furthermore, with a margin-based loss function LDAM, the improvement is much enlarged.

\end{document}